\documentclass[journal]{IEEEtran} 
\usepackage{times}

\usepackage{cite}
\usepackage{multicol}
\usepackage[bookmarks=true]{hyperref}
\usepackage[utf8]{inputenc}

\usepackage{xcolor}
\usepackage{etoolbox}
\usepackage[linesnumbered,vlined,ruled,commentsnumbered]{algorithm2e}
\SetKwRepeat{Do}{do}{while}

\SetVlineSkip{0.2em}

\SetInd{0.2em}{0.7em}

\DontPrintSemicolon{}

\SetKwIF{If}{ElseIf}{Else}{if}{}{else if}{else}{endif}

\SetAlgoSkip{}

\SetAlCapHSkip{0cm}

\SetAlgoCaptionLayout{small}
\SetAlFnt{\small}

\SetArgSty{textnormal}

\SetKwComment{Comment}{$\triangleright$\ }{}

\makeatletter
\patchcmd\algocf@Vline{\vrule}{\vrule \kern-0.4pt}{}{}
\patchcmd\algocf@Vsline{\vrule}{\vrule \kern-0.4pt}{}{}
\makeatother

\usepackage{graphicx}
\usepackage{subcaption}

\usepackage{amsmath}
\usepackage{amssymb}
\usepackage{amsfonts}
\usepackage{amsthm}
\usepackage{amsbsy}
\usepackage{mathrsfs}
\usepackage{mathtools}

\usepackage{tikz}
\usetikzlibrary{shapes,arrows}
\usetikzlibrary{calc}

\usepackage{multirow}
\usepackage{siunitx}
\usepackage{booktabs}

\usepackage[capitalise]{cleveref}

\usepackage{enumerate}

\usepackage[switch]{lineno}

\usepackage{stackengine}
\setstackgap{S}{2pt}

\newcommand*\diff{\mathop{}\!\mathrm{d}}

\newcommand{\rb}[1]{#1}

\title{\Huge Long-Horizon Multi-Robot Rearrangement Planning for Construction Assembly}

\author{Valentin N. Hartmann$^{1,2}$, Andreas Orthey$^{2}$, Danny Driess$^{2}$, Ozgur S. Oguz$^{1,3}$, Marc Toussaint$^{2}$
\thanks{This research has been supported by the Deutsche Forschungsgemeinschaft (DFG, German Research Foundation) under Germany's Excellence Strategy -- EXC 2120/1 -- 390831618.}%
\thanks{$^{1}$Machine Learning \& Robotics Lab, University of Stuttgart, Germany
        {\tt\footnotesize \{firstname\}.\{lastname\}@ipvs.uni-stuttgart.de}}%
\thanks{$^{2}$Learning and Intelligent Systems Group, TU Berlin, Germany}%
\thanks{$^{3}$Department of Computer Engineering, Bilkent University, Turkey}%
}

\IEEEoverridecommandlockouts

\begin{document}
\maketitle
\IEEEpeerreviewmaketitle

\begin{abstract}
Robotic assembly planning enables architects to explicitly account for the assembly process during the design phase, and enables efficient building methods that profit from the robots' different capabilities.
Previous work has addressed planning of robot assembly sequences and identifying the feasibility of architectural designs.
This paper extends previous work by enabling planning with large, heterogeneous teams of robots.

We present a planning system which enables parallelization of complex task and motion planning problems by iteratively solving smaller subproblems. 
Combining optimization methods to solve for manipulation constraints with a sampling-based bi-directional space-time path planner enables us to plan cooperative multi-robot manipulation with unknown arrival-times.
Thus, our solver allows for completing subproblems and tasks with differing timescales and synchronizes them effectively.  
We demonstrate the approach on multiple case-studies to show the robustness over long planning horizons and scalability to many objects and agents of our algorithm.
Finally, we also demonstrate the execution of the computed plans on two robot arms to showcase the feasibility in the real world.

\end{abstract}
\IEEEpeerreviewmaketitle


\section{Introduction}

\IEEEPARstart{A}{s robots} become ubiquitous in manufacturing and production processes, more robust algorithms to coordinate their work are needed. 
When multiple robots are employed to achieve a desired goal, two main problems have to be solved: (\textit{i}) assigning tasks to individual robots, and (\textit{ii}) coordinating movements of robots to allow effective execution of those tasks.
Combined task and motion planning (TAMP) approaches provide a suitable framework to jointly solve such problems.
Scaling these methods to robotic teams consisting of multiple agents and to long-horizon problems remains a major challenge.

We focus on multi-robot planning problems in the context of building construction:
As one of the largest industries worldwide, building construction can benefit from autonomous robots and planning processes~\cite{mckinsey2020}, and the increased efficiency they bring.
While robotic construction processes have gained more use in this industry~\cite{delgado2019,Wagner20}, particularly in off- and on-site prefabrication, an integrated autonomous decision-making and robot motion planning approach is missing.\looseness=-1
 
Previous work on autonomous assembly planning showed promising results in problems such as furniture assembly~\cite{knepper2013ikea,rodrigurez2020ral,suarez2018can} or building construction~\cite{Garrett2020RSS, Hartmann2020, leder_distributed_2019}.
Although TAMP formulations are theoretically suitable for such problems, existing approaches do not scale well with an increasing number of robots~\cite{Hartmann2020, Garrett2020RSS} and/or are only demonstrated on problems spanning short time-horizons~\cite{Toussaint2018,pan2021general}.

\begin{figure}[t]
    \centering
    \includegraphics[width=.97\linewidth]{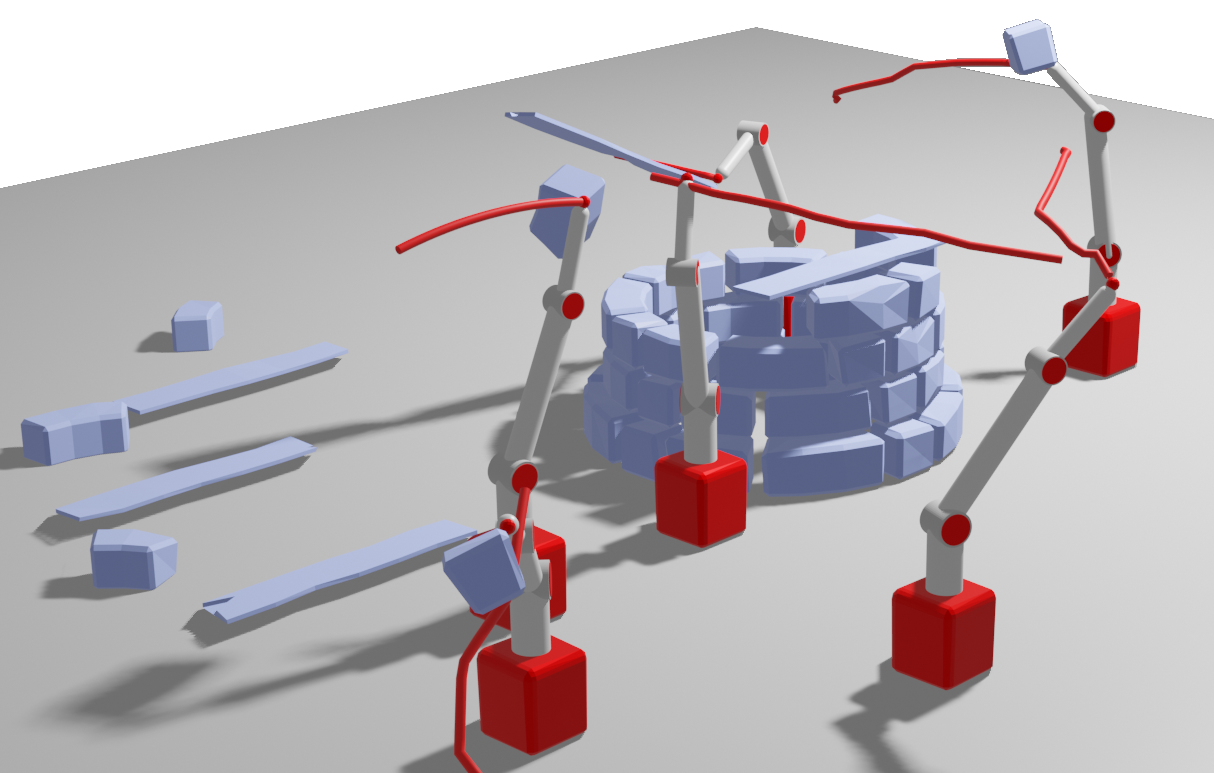} 
    \caption{
    Six mobile manipulators (red) assembling a well (blue). 
    We visualize the end-effector path of each robot as a red curve.}
    \label{fig:example_image}
\end{figure}

(Construction) Assembly planning can be thought of as rearrangement planning \cite{ota2002, stilman2007manipulation} with additional difficulties:
\textit{(i)} The ordering of objects is crucial to finish a task, since the feasibility to place a part is highly dependent on previously placed objects~\cite{Erdmann1987, Garrett2020RSS} (e.g., for stability-reasons, or object availability). 
\textit{(ii)} Objects and object orderings might impose constraints on the robots (e.g., if an object can not be transported by a single robot or temporary support of the intermediate structure is necessary).
Using multiple robots to parallelize the process, these dependencies are amplified due to the non-sequential assembly process. 

This paper extends prior work~\cite{Garrett2020, Hartmann2020} in robotic assembly planning to provide effective coordination of potentially heterogeneous robotic teams in long-horizon settings.
We frame the problem as a TAMP problem and use the framework of logic-geometric programming (LGP) \cite{Toussaint2018} to formalize it.

Our algorithm decomposes the overall planning problem by sequentially considering smaller, limited horizon problems with only a subset of robots to enable scalability.
These subproblems aim to solve the task sequencing and motion planning while keeping previously planned robot trajectories fixed.
A heuristic prioritizes the order in which subproblems are solved. 
The path planning for one subproblem needs to account for the previously computed, concurrent motion of other robots.
We present a novel bi-directional space-time embedded RRT method that allows planning to an unknown arrival time and enables accounting for previously planned robot-trajectories.
The goals for path planning are generated using an optimization-based approach that jointly solves for all manipulation constraints within the subproblem.


Our contributions are: 
\begin{itemize}
    \item a way to \emph{decompose} the assembly problem into limited horizon subproblems for only a subset of agents, \emph{while accounting for the constraints} implied by previously solved subproblems, both in time and space,
    \item a \emph{time-embedded keyframes optimization} that samples time embeddings of manipulation constraints and employs optimization methods to jointly solve for them in the limited horizon subproblems,
    \item a novel \emph{bi-directional space-time motion planner} that finds paths between keyframes in combined space-time with unknown arrival times, thereby integrating sampling-based path planning with optimization-based methods.
\end{itemize}

We demonstrate our algorithm on various construction assembly problems with up to 12 heterogeneous agents, including two long-horizon case-studies using real architectural models where up to 113 parts have to be placed.
Our evaluations show that this decomposition-based approach scales well with the number of robots and can efficiently coordinate robotic teams. 
We also show the feasibility of the plans computed by our algorithm on real robots.
\section{Related Work}

Construction robotics \cite{Saidi2016} is a growing industry which aims to automate building construction.
Specialized robots such as bricklaying robots \cite{Fastbricks2015} or automated hydraulic excavators \cite{Johns2020} recently emerged. 
Such robots have been used in various constructions settings \cite{Augugliario2014, Mascaro2020, Naboni2021,parascho2020robotic}. 
However, the algorithmic aspects of construction planning when dealing with multiple robots have only begun to be studied. 

\subsection{Multi-Robot Motion Planning}

On the lowest level, construction planning needs to solve a multi-robot motion planning problem for heterogeneous robot teams \cite{Lavalle2006}. 
This problem is often divided into two categories: First, one can plan roadmaps in parallel for individual robots, combine those roadmaps into an (implicit) joint state space roadmap and eventually search this roadmap using algorithms such as M* \cite{Wagner2015}, or discrete RRT \cite{Shome2020, Solovey2016}. 
Those algorithms can often be significantly improved using heuristics learned from prior experience \cite{Ha2020}. 

Second, prioritization frameworks \cite{Bennewitz2001,Erdmann1987, Wu2020} can be used, where robot-paths are planned sequentially, imposing the previously planned movements as constraints for the next robot. 
Planning sequentially can require backtracking, which can be time consuming.
Several approaches exist to avoid backtracking, e.g., analyzing start or goal conflicts \cite{VanDenBerg2009}.

In our approach, we use a prioritization approach by employing a heuristic to prioritize exploration of LGP subproblems. 
Contrary to prioritization in multi-robot motion planning we embed planning in space-time \cite{hsu2002randomized, Reif1994} to enable planning in a dynamic environment and with unknown arrival times.
Space-time planning has previously been investigated in the context of rendezvous planning \cite{sintov2014time}. 
In assembly tasks, it is often necessary to sample several keyframes at different time instances, due to blockages of the goal region from other agents.
We combine space-time planning explicitly with discrete constraint switches. 
Our novel bidirectional space-time RRT allows us to efficiently find time-dependent plans while avoiding collisions with previously planned robots.

\subsection{Assembly Planning}
The algorithmic aspects of construction planning are studied in the field of assembly or rearrangement planning \cite{Halperin2000, Krontiris2015}. 
Assembly planning traditionally focuses on finding valid sequences to assemble objects \cite{de2012computer, lee92}.
More recently, this was also applied to masonry constructions \cite{Kao2017,bruun2021three}.

We focus here on integrated task and motion planning applied to robotic assembly planning.
A review of solution approaches to TAMP problems can be found in \cite{Garrett2020}.
We focus on some approaches that are especially pertinent to our work:
Constraints and their intersections are explicitly enumerated in the state space \cite{Alami1994}. 
Such a constraint graph can be exploited by biased sampling at constraint intersections \cite{Khoury2021ICRA, Vega2020}, and these samples can be connected along the constraint manifolds using projection methods \cite{Hauser2011, Kingston2019, Mirabel2018}. 
Complex applications of such an approach are demonstrated in \cite{Shome2020}, such as concurrent handovers between multiple robots with multiple objects and capacity constraints.

On the other hand, symbolic assembly approaches explicitly introduce symbolic states to find task-level decisions and to factorize the problem \cite{Kaelbling2011, Toussaint2018, Toussaint2020}. 
Once symbolic decision sequences (skeletons) are found, lower level planners are used to execute a skeleton, using sampling-based \cite{Dantam2018, Thomason2019} or optimization-based \cite{Toussaint2018} methods. 
This approach can be tailored towards many different applications, for example by including force constraints \cite{Toussaint2020}, dealing with re-planning \cite{Migimatsu2020, Schmitt2019} or handling partial observability \cite{Phiquepal2019}.

Previous work dealt with long-horizon construction planning for a single agent \cite{funk2022learn2assemble}, or two agents \cite{Hartmann2020}, but solving long-horizon TAMP problems using multiple robots is an open challenge.
Initial steps towards solving multi-robot, multi-object rearrangement tasks in a simple configuration space for homogeneous robot teams with few objects were made in \cite{levihn2012multi,pan2021general}.
Such previous work assumes that the actions of robots are synchronized \cite{Hartmann2020,Toussaint2018,Shome2020,pan2021general}, plan in the combined space of all robots, and do therefore not scale \cite{Hartmann2020}, are only demonstrated for simple robots and few objects and state that they do not expect to scale \cite{levihn2012multi}, or are not demonstrated on long time-horizons~\cite{Shome2020, pan2021general}.

With our approach, we are able to plan for heterogeneous robot teams with complex interactions, and to scale to more objects and robots.
We achieve this by combining sampling based methods for path-planning, and optimization-based methods for finding the mode-switches.
Contrary to previous work, our approach does also not assume synchronicity of actions, thereby allowing the parallelization of assembly tasks efficiently.


\section{Multi Robot Rearrangement Planning Notation and Problem Formulation:}\label{sec:LGPBackground}


Given $n$ unique objects, indexed by $o\in\mathcal{O}$, $|\mathcal{O}|=n$, with initial poses $p_o^0\in SE(3)$ at time $t = 0$, and $m$ robots, indexed by $r\in\mathcal{R}$, $|\mathcal{R}|=m$, the aim is to \emph{rearrange} all objects to their (given) goal locations $p_o^G\in SE(3)$.
Each robot may have its own configuration space $\mathcal{Q}_r\subset\mathbb{R}^{{d_r}}$.

We formulate the problem as a non-linear mathematical program over the path $x:
[0,T]\rightarrow\mathcal{X}$.
The configuration space $\mathcal{X}=\mathcal{Q}\times SE(3)^n$ consists of all robot configuration spaces
$\mathcal{Q} = \mathcal{Q}_{r_1}\times \cdots \times \mathcal{Q}_{r_m}$ and object configuration spaces.

Over time, different constraints on the path are active, e.g., at the end of a $pick$-action, the end-effector of an agent needs to fulfill gripping-constraints.
Which constraints are active is determined by the task assignment $s \in \mathcal{S} = \mathcal{S}_{r_1}\times \cdots \times \mathcal{S}_{r_m}$, where $\mathcal{S}_r$ indicates the feasible tasks for robot $r$.
Thus, the state $s$ determines the current task assignment of each robot\footnote{This is slightly different to most TAMP literature, where $s$ is a set of grounded literals that determine, e.g., which robot is assigned to which task, and STRIPS-like rules determine feasible transitions between logical states (task assignments).}. 
We use $s_{r,1:K_r}\in\mathbb{S}(\mathcal{R}, \mathcal{O})$ to denote the discrete sequence of tasks of robot $r$, with $s_{r,j}\in\mathcal{S}_r$.
$K_r$ is the number of discrete states for robot $r$ in the sequence $s_{r,1:K_r}$.
The set $\mathbb{S}(\mathcal{R}, \mathcal{O})$ denotes all valid state sequences induced by a first-order logic-language for the robots $\mathcal{R}$ and objects $\mathcal{O}$.
For example, a \emph{handover}-action necessitates a \emph{pick}-action as precondition.

In most approaches to solve TAMP problems the transitions between task assignments occur at fixed intervals~\cite{Hartmann2020,Toussaint2018,Shome2020,pan2021general}.
Our problem formulation allows for task assignments to switch (finitely often) at any time.
This is achieved by the \emph{scheduling function}, $k:[0,T]\rightarrow \mathcal{N} = \left(1,\ldots,{K_1}\right)\times \cdots \times \left(1,\ldots,{K_m}\right)$, that maps continuous time $t$ into a vector of indices that select the currently active task assignments for all robots, such that $s(t) = s_{k(t)}$.
We use $k_r(t)$ to denote the scheduling function for robot $r$.
\rb{The scheduling function is constrained to respect the order of the indices, i.e., $k_r(t_1)\leq k_r(t_2)\ \forall t_1\leq t_2$.}

\begin{figure}[t]
    \centering
    \includegraphics[width=1.\linewidth]{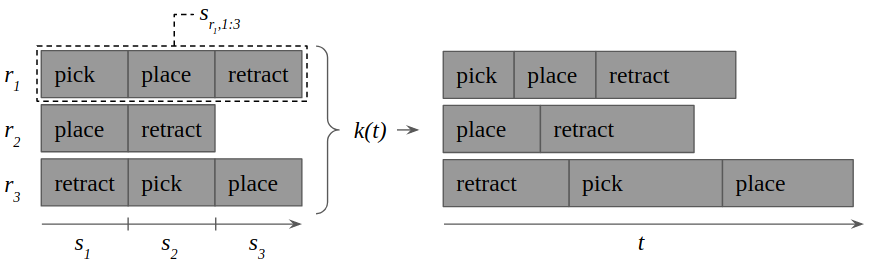} 
    \caption{
    \rb{Illustration of the scheduling function $k$ on a problem with three agents ($r_1, r_2, r_3$).
    The scheduling function maps the time $t$ to the active tasks $s_1, s_2, s_3$.}
    }
    \label{fig:sequence_illustration}
\end{figure}

Therefore, we try to find the path $x$, the terminal time $T>0$, the scheduling function $k$, and the sequences of discrete states $\left\{s_{r, 1:K_r}\right\}_{r=1}^m$ to optimize
\begin{subequations}\label{eq:rearrangement}
\begin{align} 
    \min_{\substack{x, T, k,\\ \left\{s_{r, 1:K_r}\right\}_{r=1}^m}} ~&\int^T_0 c(x(t), \dot x(t), \ddot x(t))\diff t \\
    \text{s.t.}~~~~~~~ & x(0) = x_0 \\
    \forall t\in[0,T]:~ & g(x(t), \dot x(t), s_{k(t)}) \leq 0 \label{eq:rearrangement:constraint}\\
    & \left\{s_{r, 1:K_r}\right\}_{r=1}^m \in \mathbb{S}(\mathcal{R}, \mathcal{O}) \label{eq:rearrangement:logic}\\
    & g_\text{goal}\left(x(T), p^G_{\mathcal{O}}, \mathcal{O}\right)\leq 0.  \label{eq:rearrangement:goal}
\end{align}
\end{subequations}
The task assignment state $s_{r, k_r(t)} \in\mathcal{S}_r$ determines currently active constraints for each robot $r$ on the path $x$ at time $t$ via the constraint function $g$ in \eqref{eq:rearrangement:constraint}. 
For example, $s_{r, k_r(t_1)} = s_{r, j}$ could specify the necessary constraints for robot $r$ such that it grasps an object at time $t_1$ as the $j$-th discrete state $s_{r, j}$ of the sequence $s_{r, 1:K_r}$.
Additionally, \eqref{eq:rearrangement:constraint} could describe collision constrains, or joint-limits.
In the following, we refer to the discontinuities in $s$ as \textit{mode-switches}, or \textit{keyframes}. 
\Cref{fig:sequence_illustration} illustrates the components of $s$, and how the sequence is mapped to continuous time.

The goal constraint \eqref{eq:rearrangement:goal} specifies that at the end, all objects have to be at their target poses.
The initial condition $x_0$ contains the initial configuration of all robots as well as the initial poses of the objects.
Finally, $c$ is a cost function, such as path length, control cost, or minimal time.
In case we are only interested in finding a \emph{feasible} solution, $c=0$.

\subsection{Assumptions}
We summarize the assumptions we made in the problem formulation:
\begin{itemize}
    \item Known initial and final position of all objects, \rb{and availability of a method to sample configurations for manipulating them.}
    \item Monotonic rearrangement: While it is possible to handle nonmonotonicity in the logic search, we assume in this work that each object is handled a single time. 
    However, we consider re-grasping of objects such as handovers.
    \item No force and torque constraints for the robots: In this work, we assume that the parts are light compared to the allowable robot-payload.
    Consequently, every object can be manipulated by a single robot under this assumption.
\end{itemize}


\section{Method}\label{sec:meth}

\tikzstyle{block} = [rectangle, draw, text width=10em, text centered, rounded corners, minimum height=2em]
\tikzstyle{line} = [draw, -latex']
\tikzstyle{proc} = [diamond, draw, aspect=2, text centered, text width=5em]

\begin{figure*}[t]
    \begin{center}
    \scalebox{.8}{
    \begin{tikzpicture}[node distance = 4.5cm, auto, every node/.style={scale=1}]
        \node [block] (prio) {Choose Subproblem (\cref{sec:tree})};
        \node [block, right of=prio] (skeleton) {Construct Keyframes Optimization Problem (\cref{sec:scheduling} \& \ref{sec:subgoal})};
        \node [block, right of=skeleton] (sample) {Solve Time-Embedded Keyframes Optimization-Problem};
        \node [block, right of=sample] (plan) {Space-Time Planning to Connect Keyframes (\cref{sec:planning})};
        
        \path [line] (prio) -- ($(skeleton.west) +(-.2, 0)$);
        \path [line] (sample.north) |- +(.2,.3) -| (plan.north);
        \path [line] (skeleton) -- (sample);
        \path [line] ($(plan.south) +(-.0, 0)$) |- +(-.2,-.3) -| (sample);
        \path [line] ($(sample.south)+(0.0,-0.5)$) |- +(-.2,-.2) -| (prio);
        
        \draw[black,thick,fill=gray, rounded corners, opacity=0.1] ($(skeleton.north west)+(-0.2,0.5)$) rectangle ($(plan.south east)+(0.2,-0.5)$);
        \draw[black,thick, rounded corners] ($(skeleton.north west)+(-0.2,0.5)$) rectangle ($(plan.south east)+(0.2,-0.5)$);
        \node[above] at ($(sample.north)+(0.0,+0.5)$){{\color{gray}Solve Subproblem}};

    \end{tikzpicture}
    }
    \end{center}
    \caption{
    A high-level description of the steps of our method. 
    \rb{The subproblem to solve (i.e. with which agent which part should be placed) is chosen using a heuristic.
    This subproblem is then solved (gray rectangle) by first finding a set of feasible keyframes that fulfill the constraints for the mode switches.
    Since other agents might already move on a previously planned trajectory, the method then finds a time-embedded path by repeatedly generating new keyframes with different time-embeddings and attempting to connect the keyframes to each other.
    Please refer to \cref{sec:meth} for a more elaborate description of each step.}
    \label{fig:overview}}
\end{figure*}
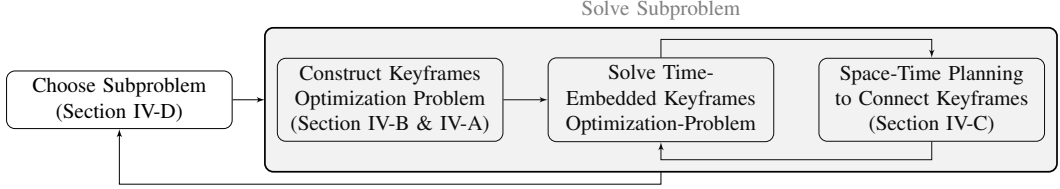


Solving the problem described in \cref{eq:rearrangement} in a fully joint and global manner is intractable, and even finding a feasible solution that utilizes all robots is hard.
We present our approach to decompose the problem into simpler subproblems, and to solve the subproblems such that the solutions together are a feasible solution to the original problem.
\Cref{alg:main,alg:solve,alg:st_rrt} and the following sections describe
\begin{enumerate}
    \item A decomposition of the overall problem into subproblems that each contain only a subset of robots and assigned tasks. 
    These subproblems account for the time-embedding in a scene where other robots are already moving, and represent coordination constraints for their respective subset of robots to ensure feasible cooperative manipulations, e.g., handovers.

    \item An approach to generate solutions to manipulation constraints defined by the subproblems, such as pick, place, or handover constraints. 
    We use this method to sample goals for the path planning algorithm.

    \item A bi-directional space-time RRT path planner to find feasible motions between keyframes taking moving robots into account. 
    
    \item A heuristic to prioritize the order of subproblems that the overall system tries to solve, and how everything is integrated with each other.
\end{enumerate}

An illustration of this system can be seen in \cref{fig:overview}.

\begin{algorithm}[t]
\caption{\rb{\texttt{plan}()}.} 
\label{alg:main}
    tree $\gets\emptyset$, curr\_node $\gets\emptyset$\;
    \While{true}{
        $\mathcal{O}\gets$ \texttt{extract\_objects}(curr\_node)\;
        $\text{seq}_\text{prev}, R_\text{prev}\gets$ \texttt{extract\_prev\_attempts}(curr\_node)\;
        $(o, R, \text{seq})\gets$ \texttt{choose\_subproblem}($\mathcal{O}, \text{seq}_\text{prev}, R_\text{prev}$)\;
        
        \If{$(o, R, \text{seq})=\emptyset$}{
            curr\_node $\gets$ \texttt{backtrack(}tree\texttt{)}\;
            \textbf{continue}\;
        }
        
        sol$_\text{sub}\gets$ \texttt{solve\_subproblem}($(o, R, \text{seq})$)\;
        \uIf{sol$_\text{sub}=\emptyset$}{
            \texttt{mark\_as\_infeasible}(curr\_node, $(o, R, \text{seq})$)\;
        }
        \Else{
            curr\_node $\gets$ \texttt{add\_to\_sol}(tree, sol$_\text{sub}$)\;
        }
        
        \If{done}{
            \Return \texttt{extract\_sol}(curr\_node)\;
        }
    }
    
    \Return Infeasible
\end{algorithm}


\begin{algorithm}[t]
\caption{\rb{\texttt{solve\_subproblem}($(o, R, \text{seq})$).}} 
\label{alg:solve}
    \For{$k=1\ldots |\text{seq}|$}{
        \texttt{goal\_sampler} $\gets$ \texttt{make\_goal\_sampler}(seq, $R, \{x_1, \ldots, x_k\}$)\;
        $q_0, t_0 \gets$ \texttt{get\_available\_time}($R$)\;
        $x_k\gets$ \texttt{ST-RRT*}($(q_0, t_0)$, \texttt{goal\_sampler})\;
        \If{$x_k = \emptyset$}{
            \Return $\emptyset$
        }
    }
    
    \For{$k=1\ldots |\text{seq}|$}{
        $x_k\gets$ \texttt{shortcut}($x_k$)\;
        $x_k\gets$ \texttt{smooth}($x_k$)\;
    }
    
    \Return $\{x_1, \ldots, x_{|\text{seq}|}\}$
\end{algorithm}

\begin{algorithm}[t]
\caption{\rb{\texttt{ST-RRT}($x_0$, \texttt{goal\_sampler})}.} 
\label{alg:st_rrt}
    $T_a\gets x_0, \ T_b\gets\emptyset$\;
    \While{not stopped}{
        $t_\text{lb}, t_\text{ub}\gets$\texttt{update\_bounds}()\;
        \If{$\text{rnd}(0, 1)< p_\text{goal}$}{
            $t\gets$ \texttt{sample}($t_\text{lb}, t_\text{ub}$)\;
            $q_g\gets$ \texttt{goal\_sampler}($t$)\;
            \texttt{add\_goal}($(t, q_g)$)\;
        }
        $q\gets$ \texttt{sample\_valid\_state}()\;
        $t\gets$ \texttt{sample\_valid\_time}($q$)\;
        \If{\textbf{not} $x_\text{new}\gets$ \texttt{extend}($(t, q), T_a$) = trapped}{
            \If{\texttt{connect}($x_\text{new}, T_b$) $=$ reached}{
                \Return \texttt{extract\_path}()\;
            }
        }
        \texttt{swap}($T_a, T_b$)\;
    }
    
    \Return $\emptyset$
    
\end{algorithm}

\subsection{Decomposition into Time-Embedded, Limited Horizon Subproblems with a Subset of Agents}\label{sec:subgoal}

A natural decomposition of \cref{eq:rearrangement} into smaller subproblems emerges from the problem specification of rearranging objects, i.e., we consider subgoals of rearranging \emph{one} object with potentially multiple robots.
The following description focuses on clarifying the degrees-of-freedom (DoF) for each subproblem. 

Assume that we are in step $l$ of the planning process.
The set $\mathcal{O}^{l-1}\subseteq\mathcal{O}$ denotes all objects that have been successfully moved to their respective goal locations at previous planning steps, i.e., an action sequence and corresponding trajectory has been planned.
The set $\bar{\mathcal{O}}^{l-1} = \mathcal{O}\backslash \mathcal{O}^{l-1}$ denotes the objects that have no plan associated yet.
A heuristic (explained in \cref{sec:tree}) selects a single new object $o_l\in\bar{\mathcal{O}}^{l-1}$ and a set of robots $R_l \subseteq \mathcal{R}$ that should be involved in rearranging the object $o_l$ to its target pose $p_{o_l}^G$.

\begin{figure}[t]
    \centering
    \includegraphics[width=1.\linewidth]{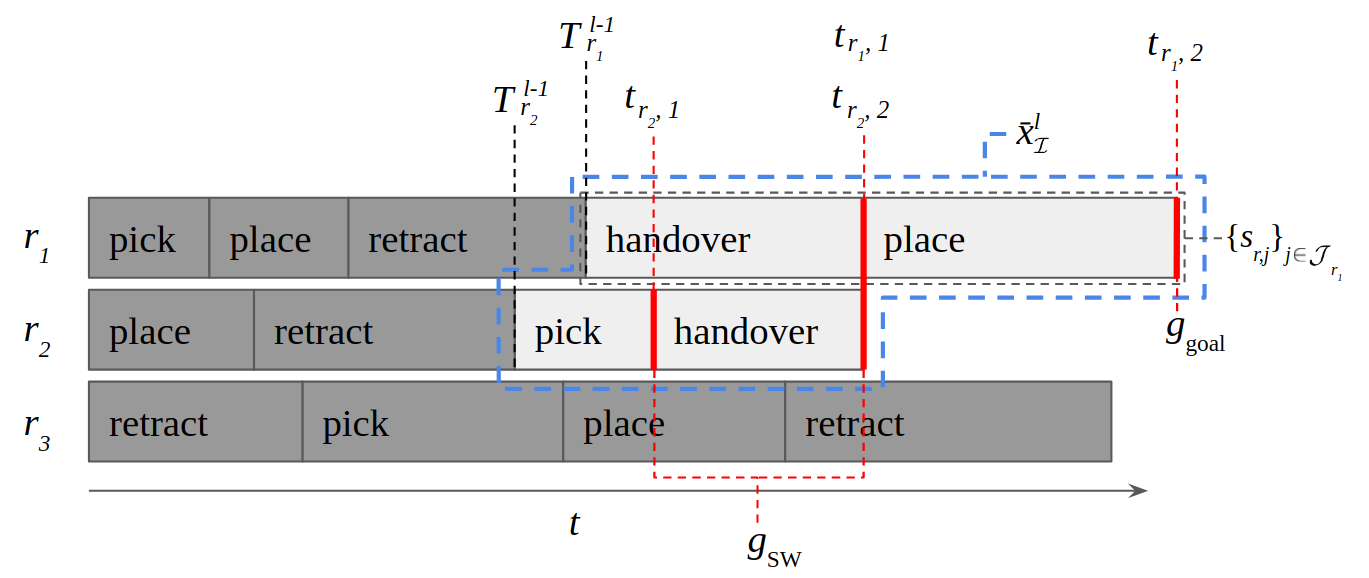} 
    \caption{
    \rb{A problem with three agents, in which we plan for $r_1$, and $r_2$ (and object $o$) with a \textit{handover} sequence. 
    The previously fixed plans are dark grey, the red lines indicates constraints ($g_\text{sw}, g_\text{goal}$) that have to be fulfilled at the mode-switches, and the actions and corresponding paths that have to be planned are light-grey.
    The blue box indicates the DoF $\bar{x}_{\mathcal{I}_l}$ for the current planning problem. 
    The active indices $\mathcal{I}$ are $\{r_1, r_2, o\}$.}
    }
    \label{fig:receding_schedule_illustration}
\end{figure}

The optimization problem we solve in step $l$ therefore only optimizes over a part of the path $x$.
We use $\bar x_{\mathcal{I}_l}^l$ to denote the degrees of freedom in the subproblem, where $\mathcal{I}_l = \left\{R_l, o_l\right\}$ is the set of indices of the path $x$ that correspond to the robots $R_l$ and the object $o_l$.
Not all degrees of freedom in $\bar x_{\mathcal{I}_l}$ necessarily become active at the same time, since some of the robots might be involved in previously planned motions up to different times (\cref{fig:receding_schedule_illustration}: $r_1$ and $r_2$ become active at different times).
Similarly, the subproblem in step $l$ is temporally embedded into a scene where robots and objects \emph{not} part of the current planning problem may follow previously computed plans ($r_3$ in \cref{fig:receding_schedule_illustration}).

To define the optimization problem for the subproblem, we therefore also need to specify how the inactive indices of $x$ are defined.
In order to do so, let $T_r^{l-1}, T_o^{l-1}\in\mathbb{R}$ denote the time until which paths for robot $r$ and object $o$ have been planned in the $l-1$ previous planning steps. 
If no path has yet been planned for a robot/object, its time is set to zero.
This allows us to define the $i$-th component of the path variable $x^l$ in the planning step $l$ at time $t$ as
\begin{align}\label{eq:dofs}
    x^l_i(t) = \begin{cases}
                x^{l-1}_i(t) & t \le T_{i}^{l-1}\\
                x^{l-1}_i\big(T_{i}^{l-1}\big) & t > T_{i}^{l-1} , ~ i \notin \mathcal{I}_l\\
                \bar{x}^{l}_i(t) & t > T_{i}^{l-1}, ~ i \in \mathcal{I}_l\\
             \end{cases}.
\end{align}
Therefore, the degrees-of-freedom $\bar x_{\mathcal{I}_l}^l$ in planning step $l$ are those of $R_l$ and $o_l$ from the point in time where they have no associated planned trajectory yet (i.e., $t>T_{i}^{l-1}, ~ i \in \mathcal{I}_l$).
In the other cases, they move according to previously computed plans ($t \le T_{i}^{l-1}$) or remain at the last planned configuration ($t > T_{i}^{l-1}$, $i \notin \mathcal{I}_l$), i.e., they correspond to inactive degrees of freedom.
The same holds for the scheduling function $k$ which has the effective degrees of freedom $\bar k_{\mathcal{I}_l}$.
Similarly, the search over the symbolic task state for the selected robots happens over $\big\{s_{r, K^{l-1}_r+1:K^l_r}\big\}_{r\in R_l}$ only.
The complete task state sequence of robot $r$ is then the concatenation $s_{r,1:K^l_r} = \left(s_{r,1}, \ldots, s_{r,K^{l-1}_r}, s_{r,K^{l-1}_r+1}, \ldots, s_{r, K^l_r}\right)$ of the sequences that have been determined in the steps up until step $l-1$ and the new sequence.
We show an illustration that serves to explain the free and fixed parts respectively in planning step $l$ in \cref{fig:receding_schedule_illustration}.

This leads to the following limited horizon optimization problem in step $l$ for the chosen object $o_l$ and robots $R_l$
\begin{subequations} \label{eq:our_rearrangement}
\begin{align} 
    \min_{\substack{\bar{T}^l,~\bar{x}^l_{\mathcal{I}_l}(\cdot),~\bar{k}^l_{\mathcal{I}_l}(\cdot) \\ \big\{s_{r, K^{l-1}_r+1:K^l_r}\big\}_{r\in R_l} }} ~&\int^{\bar{T}^l}_{\bar{T}^{l-1}} c\left(x^l(t), \dot x^l(t), \ddot x^l(t)\right)\diff t \\
    \text{s.t.}~~~~~~~~~~~ & x^l \text{ as defined in \cref{eq:dofs}} \notag \\
    \forall t\in\left[\bar{T}^{l-1},\bar{T}^{l}\right]:~\ & g\left(x^l(t), \dot x^l(t), s^l_{k(t)}\right) \leq 0\label{eq:our_rearrangement:constr}\\
    \forall r\in R_l:~\ & s_{r, 1:K^l_r} \in \mathbb{S}(R_l, \{o_l\}) \label{eq:our_rearrangement:logic}\\
    & g_\text{goal}\left(x^l(\bar{T}^l), p_{o_l}^G, \{o_l\}\right)\leq 0.  \label{eq:our_rearrangement:goal}
\end{align}
\end{subequations}
Here, $\bar{T}^{l-1}=\min_{r \in R_l} T^{l-1}_{r}$ is the earliest time for which no plan of a robot in $R_l$ exists yet.
If $R_l$ contains more than one robot, the final time $\bar{T}^l$ that is being optimized for is the maximum time of all robots $R_l$ that are involved in the current planning step, as one robot could fulfill all its constraints earlier than the others.
Consequently, the final times $T^{l}_j\le\bar{T}^l$ are assigned by extracting the minimum times where each individual robot and object $j\in\mathcal{I}_l$ fulfill their constraints.


\subsection{Time-Embedded Keyframes Optimization to Jointly Solve for Sequential Transition Constraints}\label{sec:scheduling}

\Cref{eq:our_rearrangement} is nonconvex due to collision avoidance, manipulation constraints, the time-embedding, and the discrete action-sequence.
To robustly find \emph{feasible} solutions to \cref{eq:our_rearrangement}, we combine a search for a valid sequence with nonlinear optimization and a sampling-based planner.
The optimizer solves for configurations at the transition between two task assignment states, while the motion planner, described in \cref{sec:planning}, iteratively finds paths between the keyframes.

Assume we are given robots $R$ and the sequences of discrete states $\left\{s^l_{r, j}\right\}_{j\in\mathcal{J}^l_r}\ \forall r\in R$ with $\mathcal{J}_r^l = \left\{K_r^{l-1}+1, \ldots, K_r^l\right\}$.
%
The problem
\begin{subequations}\label{eq:skeletonsampler}
\begin{align} 
\min_{x^l_{\mathcal{I}_l}(\cdot)}~~ & \sum_{\substack{r\in R_l,\\ j\in\mathcal{J}_r}} c_d(x^l_{\mathcal{I}_l}(t_{r,j}))  \\	
\text{s.t.~}~\forall r\in R_l ~\forall j\in\mathcal{J}_r:~ & g_\text{sw}(x^l_{\mathcal{I}_l}(t_{r,j}), s_{r, j}, s_{r, j-1}) \leq 0\label{eq:skeletonsampler:constraints}\\
\exists r\in R_l \exists j \in\mathcal{J}_r:~ & g_\text{goal}(x^l_{\mathcal{I}_l}(t_{r,j}), p_{o_l}^G, o_l) \le 0
\end{align}
\end{subequations}
thus describes the configurations of the involved robots and the object $o_l$ at the mode switching times $t_{r,j}$.
The constraint \eqref{eq:skeletonsampler:constraints} is the discrete version of \eqref{eq:our_rearrangement:constr} at the transition from $s_{r,j-1}$ to $s_{r,j}$.
The cost function $c_d$ is the discrete version of $c$. 
Solving \cref{eq:skeletonsampler} (using, e.g., \cite{deharo2020learning}) generates a set of keyframes that jointly fulfill interdependent  constraints, e.g., finding consistent hand positions in pick and place poses.

To solve \cref{eq:skeletonsampler}, we first \rb{uniformly} sample the times $t_{r,j}$ with $t_{r,j-1}< t_{r,j}$ and $t_{r,j} > \bar{T}^{l-1}$ for $j\in\mathcal{J}_r$ where the transition from the discrete state $s_{r,j-1}$ to $s_{r,j}$ occurs.
We then use an optimization based solver to find configurations fulfilling the constraints.
Optimization-based solvers are strong to resolve equality constraints, but are prone to local optima. 
We alleviate this problem by \emph{repeatedly} solving \cref{eq:skeletonsampler} to generate various consistent keyframes. 
There are two reasons for why the solution to \cref{eq:skeletonsampler} is randomized and generates varieties of solutions: 
First, whenever we solve \cref{eq:skeletonsampler}, we sample the times $t_{r,j}$, which leads to a different time embedding and corresponding constraints. 
Second, we initialize the optimizer with randomly sampled configurations, which helps to find various local optima.

In this view, solving \cref{eq:skeletonsampler} generates feasible keyframes that can be used as goals for bi-directional path planning.

\subsection{Bi-directional Space-Time Path Planning to Connect Keyframes}\label{sec:planning}
\begin{figure*}[t]
    \centering
    \includegraphics[width=.9\linewidth]{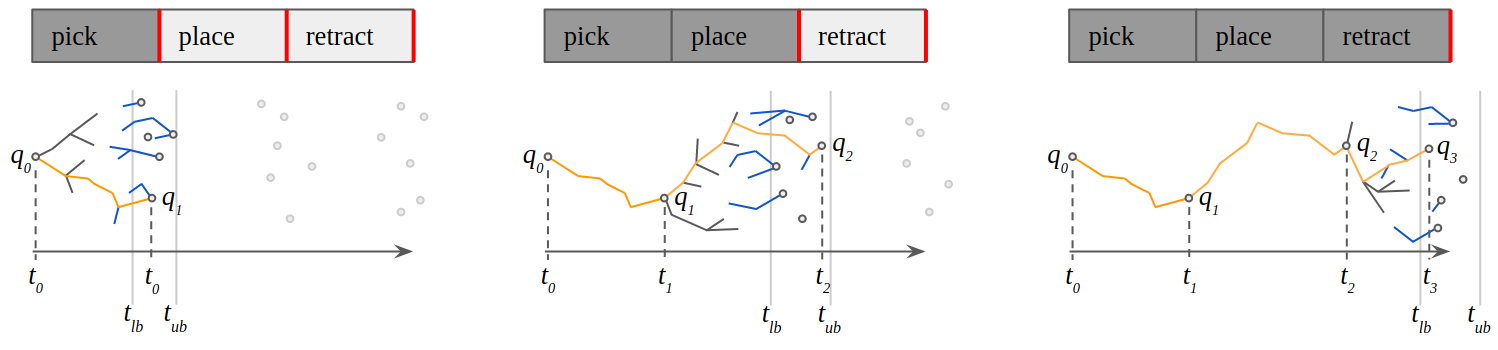} 
    \caption{
    \rb{Illustration of how plans are computed sequentially for an action sequence using the keyframe sampler and the bi-directional space-time planner.
    In the action sequence, red shows the constraints that are part of the optimization problem to generate the keyframes, dark grey indicates the actions for which a path is fixed.
    In the path planning, the nodes are the keyframes that are computed, where the light grey ones are the keyframes that are computed for the later mode-switches. The edges of the forward tree are dark grey, and the ones of the reverse tree are blue. The final path for an action is orange. 
    Please refer to \cref{sec:planning} for an in-depth explanation.}
    }
    \label{fig:strrt_illustration}
\end{figure*}

We compute the path for a given action sequence in a sequential manner (\cref{fig:strrt_illustration}): the path planner first aims to find a path between the first two keyframes; when one is found, it moves on to find a path to a third keyframe that is consistent with the first two. 
It can always query the keyframes optimizer for more keyframes consistent with given previous keyframes, i.e., \cref{eq:skeletonsampler} is solved for the \textit{remaining} mode-switch configurations. 
Solving the full remaining problem, instead of only the constraints implied by the next mode-switch, excludes keyframes that are feasible in the next mode switch, but lead to an infeasible problem later on, e.g., a \textit{pick} configuration, which does not correspond to a feasible \textit{place}-configuration.

Since the arrival time at which the robot can reach a goal keyframe configuration is unknown, we uniformly sample a range of candidate time-embeddings $t_{r,j}$ as input to the keyframes optimizer. 
During the planning process, the path planner continuously extends the range of time-embedding samples to allow for consideration of larger time-spans, i.e., to enable `waiting'.
The arrival times that are found in this fashion correspond to the scheduling function $k$ for the robots involved in the current subproblem.

\subsubsection*{Path-Planning}
We finally present the bi-directional path planner to connect the keyframe-configurations. 
Following the standard notation for time-embedded path planning \cite{Reif1994, sintov2014time}, let $\mathcal{Q}_{R_l}$ denote the configuration space for the robots $R_l$, and $\mathcal{T}\subset\mathbb{R}_{\geq 0}$ the time dimension. 
Our path planning problem is the problem of finding a collision-free path through the combined space-time $\mathcal{Y}=\mathcal{Q}_{R_l}\times \mathcal{T}$ from an initial keyframe configuration $(x(t_{r,j-1}), t_{r,j-1})$ at time $t_{r,j-1}$, to a \emph{set} of candidate goal keyframe configurations, each a pair $(x(t_{r,j}), t_{r,j})$ in space-time with varying $t_{r,j}$.







The free configuration space $\mathcal{Q}_\text{free}$ for the robots $R_l$ is time-dependent, as objects and other agents might move on previously planned paths. 
Additionally, it can be the case that some of the agents we are currently planning for move on a fixed path for some time, i.e., they only become a degree of freedom at time $t>\bar{T}^{l-1}$.
We deal with this by using a constrained path-planner \cite{berenson2009manipulation}, with the constraints defined in \cref{eq:dofs}.

%
%

Specific care has to be taken due to the time-dimension, and the direction-dependent distance function in the configuration space:
\begin{equation}
\arraycolsep=2pt\def\arraystretch{1}
    d(y_1, y_2)\! =
    \begin{cases}
    \lambda d_{\mathcal{Q}_{R_l}}(q_1, q_2)\! +\! (1\!-\!\lambda)(t_2\!-\!t_1),\\
    &\phantom{\text{(Oo}}\llap{\text{ if }$t_1<t_2, ~v\leq v_{\text{max}}$;} \\
    \infty, &\text{else.}
    \end{cases}
\end{equation}
where $d_{\mathcal{Q}_{R_l}}$ can be any valid metric in $\mathcal{Q}_{R_l}$, and $v$ is an estimate of the velocity. 
We use $\lambda\in(0,1]$ to describe the importance of the path-length and the needed time respectively.
This distance-metric encodes the inability to move from a configuration $q_1$ at time $t_1$ to a configuration $q_2$ at time $t_2$ if either the required speed $v$ is too high, or the robot would need to move backwards in time.

We then extend bi-directional rapidly-exploring random trees (RRT)~\cite{Kuffner2000} to space-time\footnote{In this work, we used an early version of the algorithm presented in~\cite{grothe2022strrt}.}:
This extension consists of the previously described keyframe sampler which generates the goals, time dependent collision queries, and configuration-sampling bounded by the sampled keyframe time. 
Specific care has to be taken when connecting edges using the goal-centered tree: we move `backwards' in this case, and thus the distance function has to be adapted.
This bi-directional space-time RRT formulation allows us to efficiently find paths. 

After finding a path using the outlined approach, we post-process the path by shortcutting~\cite{Geraerts2007} the obtained path, and smoothing it.
\rb{
Shortcutting of the path works by repeatedly choosing two discrete states of the path and checking if they can be connected with a straight line while fulfilling the constraints, e.g., collision, kinematics, or velocity limits.
If that is the case, the straight line path replaces the part of the path between the chosen discrete states.
}

\rb{
For smoothing, we use an optimizer~\cite{14-toussaint-KOMO} that takes the constraints and the cost function of the original problem as input and is thus able to take the dynamics and constraints into account.
Taking this approach is similar to separating path planning and trajectory planning, and is a common approach to find feasible paths in cluttered environments \cite{kant1986toward,pham2017admissible}.
}

As in the planning itself, care has to be taken to shortcut and smooth in a constrained manner, i.e., parts of the path that have previously been fixed can not be altered.

\begin{figure*}[t]
\centering
    \begin{subfigure}[t]{.25\textwidth}
        \centering
        \includegraphics[width=.97\linewidth, height=0.8\linewidth]{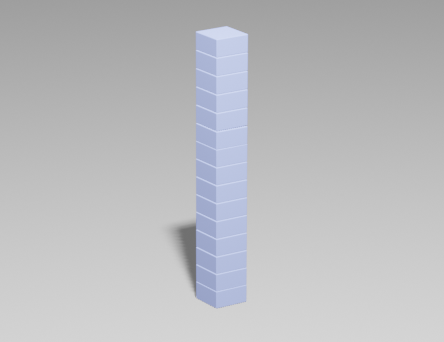} 
        \caption{Tower}
    \end{subfigure}%
    \begin{subfigure}[t]{.25\textwidth}
        \centering
        \includegraphics[width=.97\linewidth, height=0.8\linewidth]{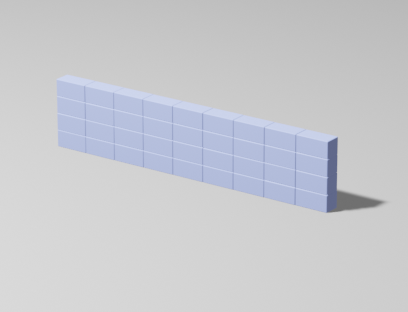}
        \caption{Wall}

    \end{subfigure}
    \begin{subfigure}[t]{.25\textwidth}
        \centering
        \includegraphics[width=.97\linewidth, height=0.8\linewidth]{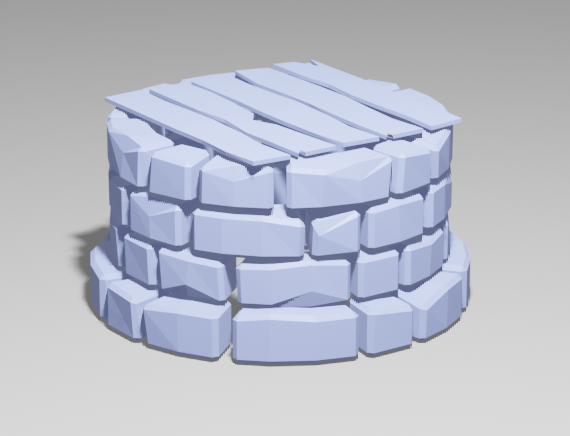}
        \caption{Well}
    \end{subfigure}%
    \begin{subfigure}[t]{.25\textwidth}
        \centering
        \includegraphics[width=.97\linewidth, height=0.8\linewidth]{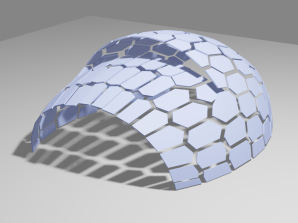}
        \caption{Pavilion}
    \end{subfigure}
    \caption{Final configurations of the models we use for the experiments and demonstrations.}
    \label{fig:models}
\end{figure*}

\subsection{Prioritization of Subproblems and Search over Skeletons}\label{sec:tree}
In each step $l$, the limited horizon formulation \cref{eq:our_rearrangement} requires a selection of object $o_l$ and robots $R_l$.
%
Due to the assembly tasks we consider, the order in which objects have to be rearranged (in particular being placed) is not obvious.
For example, objects need sufficient support in order to be placed, while placing some objects too early might obstruct later object placements.

In planning step $l$ we have the set $\mathcal{O}^{l-1}\subseteq\mathcal{O}$ of previously rearranged objects, and the set of objects $\bar{\mathcal{O}}^{l-1} = \mathcal{O}\backslash \mathcal{O}^{l-1}$ that have no plan associated yet. 
From previous task assignments and trajectory planning, the times $T^{l-1}_r$ until which a trajectory is already assigned to robot $r$ are known.
Based on this planning state, the algorithm has to 
\begin{enumerate}[(i)]
    \item choose an object $o_l\in\bar{\mathcal{O}}^{l-1}$ subject to object order constraints (expressed  by $\phi(o_l; \mathcal{O}^{l-1})\le 0$),
    \item choose a subset of robots $R_l\subseteq \mathcal{R}$ to rearrange object $o_l$,
    \item search through the space of assignment sequences $\big\{s_{r, K^{l-1}_r+1:K^l_r}\big\}_{r\in R_l}$ to find a sequence that is logically feasible, leads to the goal (i.e., object placement), and for which the keyframes optimizer and path planner can find solutions. If no such sequence is found, choose another robot assignment $R_l$ (that was not chosen yet).
    \item If no possible choice of $R_l$ leads to a feasible subproblem (\cref{eq:our_rearrangement}), we backtrack to rewind previous object placements, and attempt to place $o_l$.
    Backtracking is repeated until a valid solution to place $o_l$ is found. 
\end{enumerate}
%
This can be seen as a depth-first tree-search over the objects, robots, and assignment sequences. 
These steps are explained in more detail in the following:

\subsubsection{and 2) Selection of Object and Robots}
The objects $o_l$ in step 1) and robots $R_l$ in step 2) are selected by a strict prioritization: We prioritize the selection of objects by minimizing a heuristic $h$:
\begin{align}\label{eq:greedy}
 h(o_l; \mathcal{O}^{l-1}) ~~\text{s.t.}~~ \phi(o_l; \mathcal{O}^{l-1})\leq 0,
\end{align}
where the constraint  $\phi(o; \mathcal{O}^{l-1})\le 0$ is application specific.
As example, it could express that intermediate construction states need to be stable, or put limits on deflections of objects.
Objects violating the constraint are excluded from the prioritized search.

The selection of robots $R_l$ is prioritized by minimizing
\begin{align}
\max_{r\in R_l}\ T^{l-1}_r,
\end{align}
i.e., the latest busy time of all involved robots.
This assumes (very conservatively) that the work on this subproblem starts only when the last of the involved robots becomes free.
The robot-selection could also be prioritized by, e.g., minimizing the estimated finishing time of the subproblem.

%

\setcounter{subsubsection}{2}



\subsubsection{Path Planning and Action Sequence/Robot Rejection}
finding an action sequence is realized as a breadth-first search to check if there exists a sequence
\begin{equation}
    \big\{s_{r, K^{l-1}_r+1:K^l_r}\big\}_{r\in R_l}\in \mathbb{S}(R_l, \{o_l\})
\end{equation}
which is logically feasible and leads to the symbolic goal.

With \cref{eq:our_rearrangement} fully defined by the choice of $R_l$, $o_l$, and $\big\{s_{r, K^{l-1}_{r}+1:K^l_r}\big\}_{r\in R_l}$, we try to find a valid path (as detailed in \cref{sec:scheduling} and \cref{sec:planning}).
Before attempting to solve the full problem, it is possible to evaluate \textit{lower bounds}, i.e., simpler subproblems, which, if they are infeasible, guarantee that there is no solution to the full problem.
In our case, examples for this are \textit{i)} attempting to find a placement pose, and \textit{ii)} attempting to find the configurations at the mode switches.
For a more thorough description of the notion of lower bounds, we refer to \cite{17-toussaint-ICRA}.

If a problem is infeasible for a chosen $\big\{s_{r, K^{l-1}_r+1:K^l_r}\big\}_{r\in R_l}$, we first attempt to solve the problem using different $\big\{s_{r, K^{l-1}_r+1:K^l_r}\big\}_{r\in R_l}$ multiple times, and if still no feasible solution can be found, we restart from \textit{2)}, excluding infeasible task sequences and robots.

In general, the methods we use for generating keyframes or motion paths can not prove infeasibility of a specific discrete assignment $R_l$, $o_l$, $\big\{s_{r, K^{l-1}_{r}+1:K^l_r}\big\}_{r\in R_l}$ due to non-convexity of the optimization problem, or due to finite runtime of the path planning.
Hence, we keep a list of all assignments that we determined to be infeasible before, and revisit them in a deprioritized manner if still no solutions in the remaining discrete assignments can be found.
This allows us to explore more promising decisions, while still guaranteeing that a solution will be found eventually, if it exists. 
For brevity, this was left out of the algorithm.

\subsubsection{Backtracking}
At this stage, all robots $R_l$, and possible task assignments were checked.
Thus, any infeasibility at this stage must be caused by previously placed objects, since a selected part $o_l$ fulfills all the constraints to be able to be placed, i.e., $\phi(o_l; \mathcal{O}^{l-1})\le 0$, and placing additional objects can never make placement of $o_l$ feasible.
Instead, we backtrack to rewind previously placed objects until a valid solution to place object $o_l$ can be found.

\section{Demonstrations \& Results}

\rb{
We analyze the scalability of the algorithm, and how some scenarios benefit more than others from better parallelizability.
We demonstrate the robustness of the algorithm on long-horizon scenarios, show the ability to coordinate multiple robots in a scenario where a handover sequence is necessary, and a real robot experiment.
Finally, we compare the algorithm to a modified version of a classical TAMP-solver.
}

\subsection{Setup}
We test the algorithm on several construction scenarios\footnote{Simulated using \url{https://github.com/MarcToussaint/rai}}\footnote{The experiments were run on a single core of Intel(R) Core(TM) i7-8565U CPU @ 1.80GHz with 16GB RAM using Ubuntu 18.04.}:\looseness=-1
\begin{itemize}
    \item A tower, consisting of 15 pieces, where the placements of the parts have to be in strictly sequential order.
    \item A wall, consisting of 36 bricks used to analyze how well a task can be parallelized.

    \item A well, consisting of 52 pieces, used to demonstrate scalability.
    \item A pavilion consisting of 113 unique wooden cassettes, used to demonstrate scalability.
\end{itemize}

The final configuration of the models is visualized in \cref{fig:models}.
We assume that the pieces form a rigid body with the neighbouring pieces as soon as they are placed, and thus neglect both the fastening process, and the structural support that would be necessary.
We are first and foremost interested in finding a \textit{feasible} solution to \cref{eq:rearrangement}, i.e. we use $c=0$.
For the real robot experiment, we minimize the acceleration in the smoothing-step.



\subsubsection{Robots}
\begin{figure}[t]
\centering
\def\heightImages{0.8\linewidth}
    \begin{subfigure}[t]{.17\textwidth}
        \centering
        \includegraphics[height=\heightImages,width=.97\linewidth]{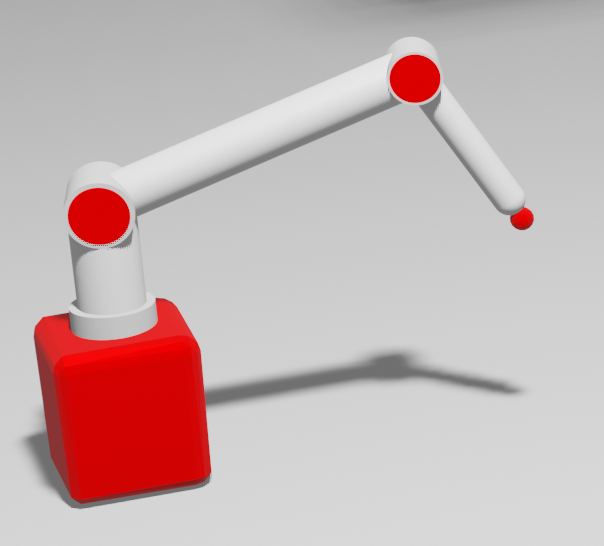} 
    \end{subfigure}%
    \begin{subfigure}[t]{.17\textwidth}
        \centering
        \includegraphics[height=\heightImages,width=.97\linewidth]{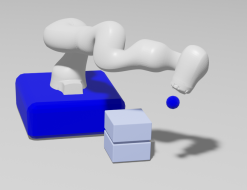} 
    \end{subfigure}%
    \begin{subfigure}[t]{.17\textwidth}
        \centering
        \includegraphics[height=\heightImages,width=.97\linewidth]{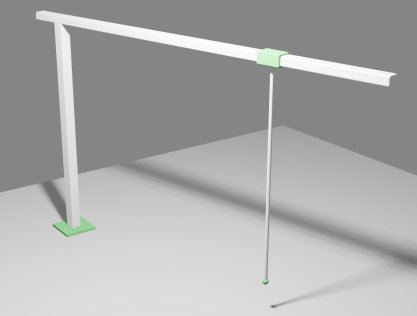} 
    \end{subfigure}
    \caption{
    An illustration of the robots (from left to right): a mobile manipulator, a KUKA-arm on a mobile base, and a crane.}
    \label{fig:robots}
\end{figure}

We demonstrate our approach using 3 different robots (\cref{fig:robots}): a mobile manipulator, a KUKA-arm on a mobile base, and a crane.
The robots we are using for the demonstrations are holonomic.
If not stated differently, we model manipulation by gripping-by-touch: On construction sites, vacuum grippers are commonly used, which can be approximated as gripping-by-touch.

\subsubsection{Task Assignments} The discrete task assignments are \textit{pick}, \textit{place}, \textit{retract} and \textit{handover}. We require a \textit{place} task to always be followed by a \textit{retract} task, since in general, after placing an object, it is not desirable to stay in the same configuration, and possibly block the placement configurations of other agents.

\subsubsection{Ordering Heuristic}
We represent the buildings as a graph, with the nodes being parts, and edges  between connected parts.
The heuristic $h$ in \cref{eq:greedy} is chosen to find the object that maximizes the number of previously placed neighbours.
The placeable parts are the set of nodes which are connected to at least one node in the graph that is already placed.
This is encoded in the constraint $\phi$ in \cref{eq:greedy}.

Due to the heuristic and placement constraint, no backtracking was required in our experiments.
As such, there is no specific demonstration, or mention of how many times backtracking was needed in the examples.

\subsection{Experimental Results}

\begin{figure}[t]
    \centering
    \includegraphics[width=.49\linewidth]{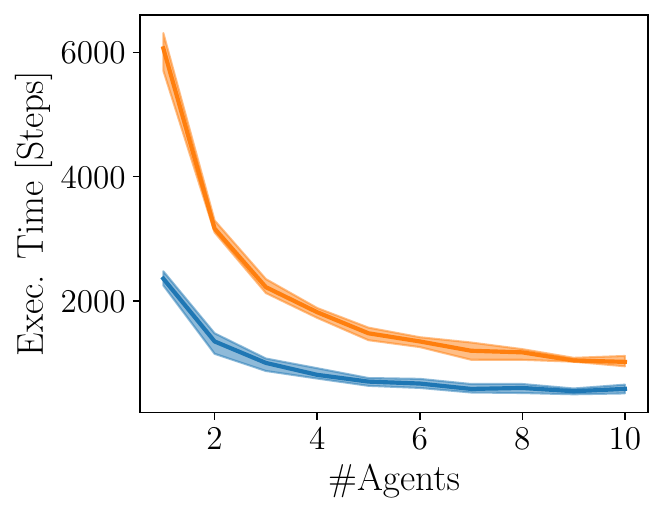}
    \includegraphics[width=.49\linewidth]{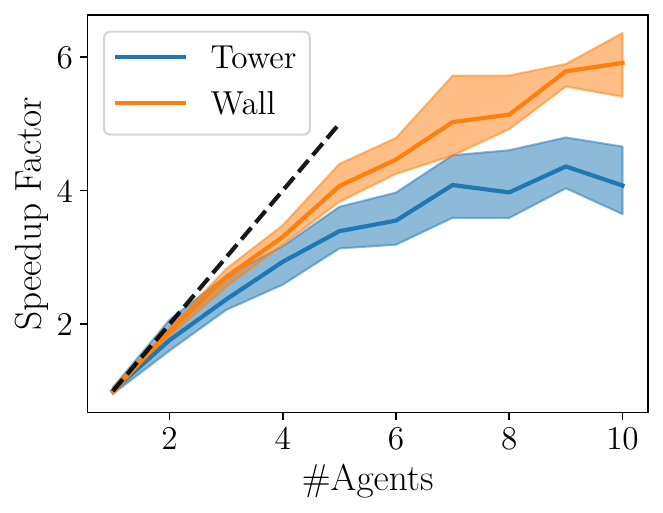}
    \caption{
    The \rb{execution time (left)}, and the speedup factor of the execution time (right) when using multiple agents for the tower and the wall models over 10 runs.
    The speedup factor is the factor by which the execution time is sped up by using $m$ agents compared to only one agent.
    \rb{The line is the median and the shaded area is bounded by the 25 and 75 percentile.}
    }
    \label{fig:agent_time_model}
\end{figure}

\begin{figure}[t]
    \centering
    \includegraphics[width=.97\linewidth]{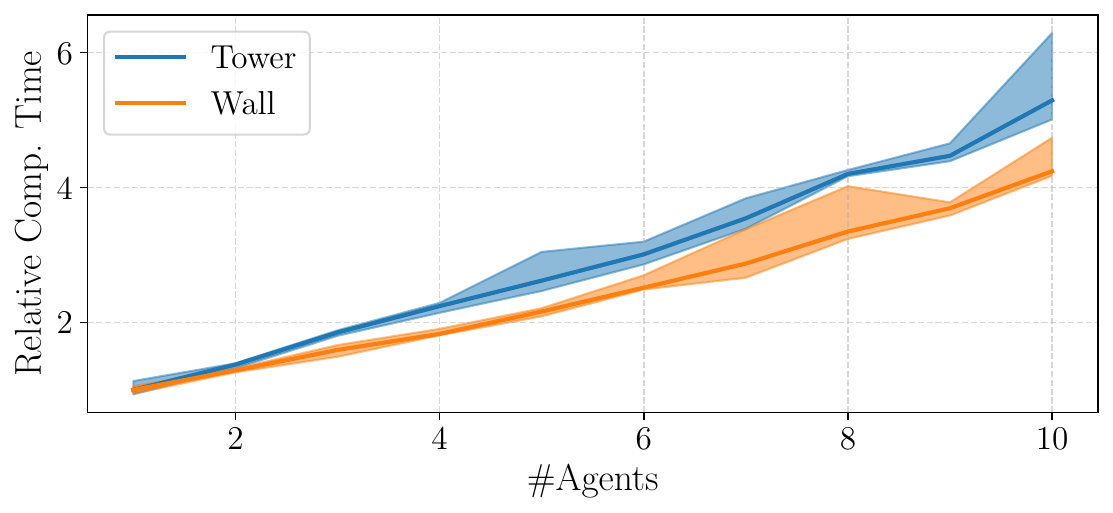} 
    \caption{
    The relative computation time for different numbers of agents for the tower and the wall scenario over 10 runs. 
    \rb{The line is the median and the shaded area is the 25 and 75 percentile, respectively.}
    }
    \label{fig:comp_time}
\end{figure}

We provide analysis on several quantitative metrics in this section.
The experiments for the analysis were done using the mobile manipulator utilizing \textit{pick and place}-sequences.
Videos of the assembly processes for various models with different team-sizes and various constellations of robot types can be found in the supplementary material.

\subsubsection{Execution Time} The execution time, i.e. the real time of the planned movements, is expected to decrease the more agents are used for the rearrangement task.
\Cref{fig:agent_time_model} plots the factor by which the execution is sped up against the number of agents for the tower and the wall, and highlights how the \textit{separability} of the model influences how adding more robots leads to diminishing returns in the speedup factor.
These diminishing returns occur more quickly for the tower, where all agents need to place the objects in close vicinity.

\subsubsection{Computation time} \cref{fig:comp_time} shows approximately linear scaling with the number of agents for the computation time.
This supports that our approach of planning each agent separately, and assuming the previously planned agents as fixed, scales well.
Since the {tower} has a bottleneck where all agents have to come together within a relatively small region, it scales worse than the {wall}.
The objects making up the wall are distributed over a larger space, and thus the agents do not impede each other as much, i.e., the planning problem is easier, since fewer agents per volume are present.
\Cref{tab:comp_time_breakdown} breaks down the average necessary computation time for the different steps of the planning process.
Specifically, we want to highlight that solving the keyframe-problem takes most of the time of the whole planning process.

\begin{table}[t]
\begin{center}
    \scriptsize
    \renewcommand\arraystretch{1.15}
    \caption{Median computation time over 10 runs for Tower and Wall, 5 for Well and Pavilion for the  components of the algorithm for all models, with number of agents $m$. 
    The total time additionally contains e.g. pre-processing of the model, deciding on action-sequences etc.
    \rb{The super-{} and subscripts are the difference to the 25, and 75 percentile.}}
    \begin{tabular}{p{9mm}p{2mm}|@{\hskip 1mm}
            S[table-format=3.1]@{\hskip 2mm}S[table-format=3.1]@{\hskip 2mm}S[table-format=3.1]@{\hskip 1mm}|@{\hskip 2mm}
            S[table-format=4.1]@{\hskip 6mm}}
    \toprule
    & & \multicolumn{4}{c}{Time [s]} \\
     & $m$ & \text{Keyframes-opt} & \text{Path-planning} & \text{Postprocessing} & \text{Total} \\
    \midrule
    \multirow{3}{*}{\shortstack[l]{Tower\\(15 obj.)}} 
                & 1 & 13.0\stackanchor{\scalebox{.6}{$+2.0$}}{\scalebox{.6}{$-1.6$}} & 0.7\stackanchor{\scalebox{.6}{$+0.1$}}{\scalebox{.6}{$-0.0$}} & 4.3\stackanchor{\scalebox{.6}{$+0.3$}}{\scalebox{.6}{$-0.2$}} & 19.3\stackanchor{\scalebox{.6}{$+2.5$}}{\scalebox{.6}{$-1.3$}} \\
                & 5 & 19.7\stackanchor{\scalebox{.6}{$+0.3$}}{\scalebox{.6}{$-1.3$}} & 4.7\stackanchor{\scalebox{.6}{$+0.4$}}{\scalebox{.6}{$-0.2$}} & 15.8\stackanchor{\scalebox{.6}{$+0.8$}}{\scalebox{.6}{$-0.6$}} & 50.6\stackanchor{\scalebox{.6}{$+8.2$}}{\scalebox{.6}{$-3.0$}} \\
                & 10 & 25.8\stackanchor{\scalebox{.6}{$+4.7$}}{\scalebox{.6}{$-1.4$}} & 13.3\stackanchor{\scalebox{.6}{$+10.9$}}{\scalebox{.6}{$-0.7$}} & 44.4\stackanchor{\scalebox{.6}{$+2.0$}}{\scalebox{.6}{$-3.6$}} & 102.3\stackanchor{\scalebox{.6}{$+19.3$}}{\scalebox{.6}{$-5.5$}} \\

    \midrule
    \multirow{3}{*}{\shortstack[l]{Wall\\ (36 obj.)}} 
                & 1 & 42.1\stackanchor{\scalebox{.6}{$+3.0$}}{\scalebox{.6}{$-2.9$}} & 4.2\stackanchor{\scalebox{.6}{$+0.2$}}{\scalebox{.6}{$-0.3$}} & 23.9\stackanchor{\scalebox{.6}{$+0.7$}}{\scalebox{.6}{$-0.8$}} & 75.4\stackanchor{\scalebox{.6}{$+4.3$}}{\scalebox{.6}{$-4.2$}} \\
                & 5 & 62.0\stackanchor{\scalebox{.6}{$+6.1$}}{\scalebox{.6}{$-0.7$}} & 17.0\stackanchor{\scalebox{.6}{$+1.1$}}{\scalebox{.6}{$-1.9$}} & 53.7\stackanchor{\scalebox{.6}{$+1.2$}}{\scalebox{.6}{$-3.1$}} & 163.0\stackanchor{\scalebox{.6}{$+3.7$}}{\scalebox{.6}{$-5.4$}} \\
                & 10 & 102.7\stackanchor{\scalebox{.6}{$+6.0$}}{\scalebox{.6}{$-4.8$}} & 50.1\stackanchor{\scalebox{.6}{$+4.2$}}{\scalebox{.6}{$-6.1$}} & 107.6\stackanchor{\scalebox{.6}{$+5.3$}}{\scalebox{.6}{$-2.4$}} & 319.7\stackanchor{\scalebox{.6}{$+37.4$}}{\scalebox{.6}{$-5.0$}} \\

    \midrule
    \multirow{2}{*}{\shortstack[l]{Well\\(52 obj.)}} 
                & 1 & 256.8\stackanchor{\scalebox{.6}{$+15.1$}}{\scalebox{.6}{$-0.7$}} & 53.1\stackanchor{\scalebox{.6}{$+11.2$}}{\scalebox{.6}{$-4.1$}} & 131.7\stackanchor{\scalebox{.6}{$+26.2$}}{\scalebox{.6}{$-3.5$}} & 450.2\stackanchor{\scalebox{.6}{$+15.5$}}{\scalebox{.6}{$-13.0$}} \\
                & 5 & 342.2\stackanchor{\scalebox{.6}{$+45.9$}}{\scalebox{.6}{$-16.1$}} & 84.1\stackanchor{\scalebox{.6}{$+11.8$}}{\scalebox{.6}{$-1.6$}} & 222.6\stackanchor{\scalebox{.6}{$+38.9$}}{\scalebox{.6}{$-1.1$}} & 722.1\stackanchor{\scalebox{.6}{$+123.3$}}{\scalebox{.6}{$-15.4$}} \\
    \midrule
    \multirow{2}{*}{\shortstack[l]{Pavilion\\ (113 obj.)}} 
                & 1 & 734.7\stackanchor{\scalebox{.6}{$+113.0$}}{\scalebox{.6}{$-38.5$}} & 311.5\stackanchor{\scalebox{.6}{$+26.5$}}{\scalebox{.6}{$-27.3$}} & 511.1\stackanchor{\scalebox{.6}{$+86.3$}}{\scalebox{.6}{$-34.2$}} & 1597.7\stackanchor{\scalebox{.6}{$+94.7$}}{\scalebox{.6}{$-37.1$}} \\
                & 5 & 937.6\stackanchor{\scalebox{.6}{$+11.5$}}{\scalebox{.6}{$-75.8$}} & 339.7\stackanchor{\scalebox{.6}{$+21.2$}}{\scalebox{.6}{$-8.0$}} & 669.0\stackanchor{\scalebox{.6}{$+36.4$}}{\scalebox{.6}{$-28.4$}} & 2336.1\stackanchor{\scalebox{.6}{$+86.2$}}{\scalebox{.6}{$-195.2$}} \\

    \bottomrule
    \end{tabular}
    \label{tab:comp_time_breakdown}
\end{center}
\end{table}

\begin{figure*}[t]
    \centering
    \includegraphics[width=.97\linewidth]{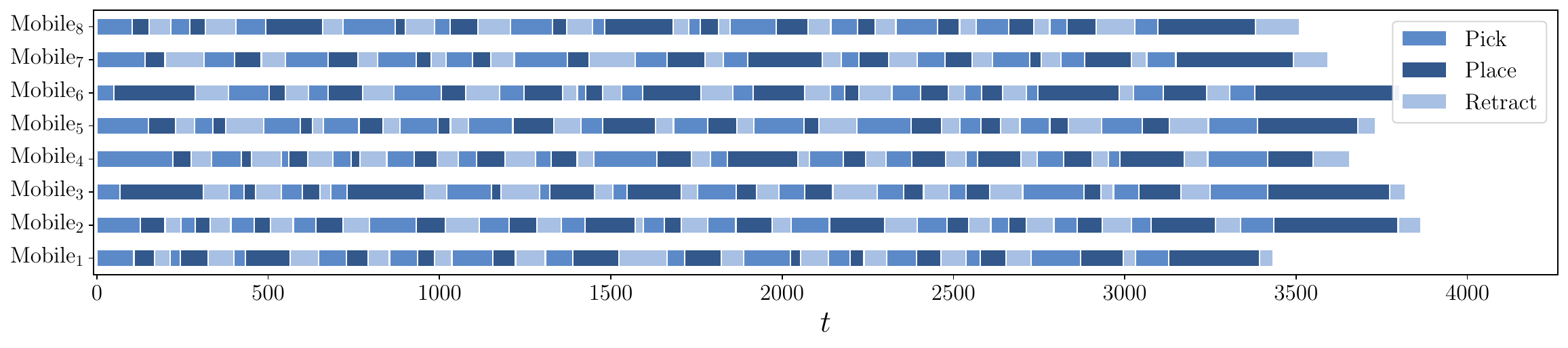} 
    \caption{The schedule for a set of 8 mobile agents assembling the pavilion.}
    \label{fig:schedule}
\end{figure*}

\subsection{Comparison to fixed-time-sampling}
We did not put an emphasis on comparison with other methods, as the other methods we are aware of (\cite{pan2021general, Shome2020, Hartmann2020, levihn2012multi, ota2002, chencooperative}) do not scale to the number of robots and objects we consider.
We compare our method to a fixed-time sampling scheme\footnote{This is similar to what most TAMP solvers would do.}, which decomposes the problem as the presented approach does, and uses prioritized planning, but does not allow for variable durations of the actions, i.e. all mode-switches take place at a multiple of $T$.
We first note that the fixed-time-sampling approach is extremely sensitive to the choice of $T$.
If there are tasks with different duration, or agents that have different maximum speeds, this approach will not allocate the right time, and take an unnecessarily long time for short tasks, or be infeasible altogether.

\begin{figure}[t]
    \centering
    \includegraphics[width=.97\linewidth]{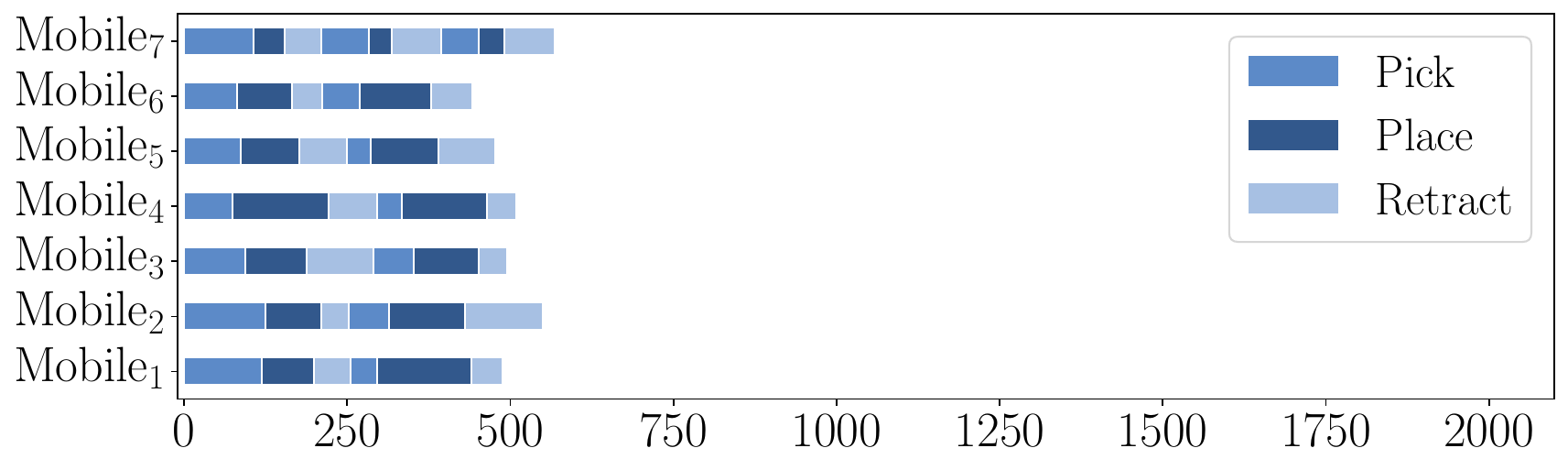} 
    \includegraphics[width=.97\linewidth]{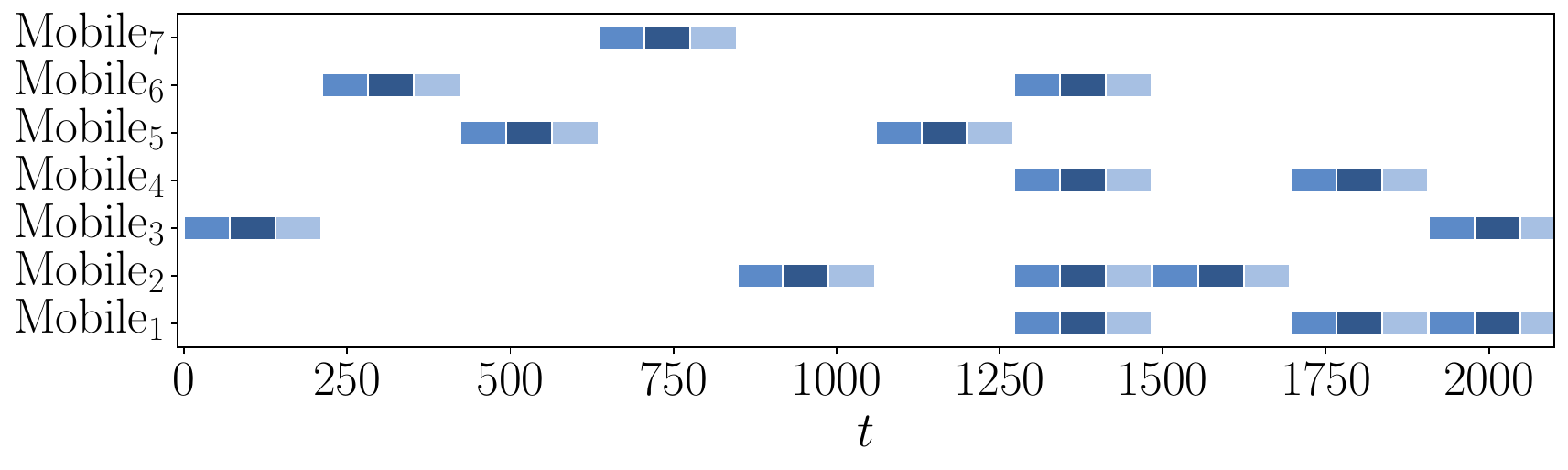} 
    \caption{
    Illustration of the schedules obtained by our method (top), and fixed-time sampling (bottom) for the tower with 7 mobile agents.}
    \label{fig:ablation_schedule}
\end{figure}

\begin{figure}[t]
    \centering
    \includegraphics[width=.97\linewidth]{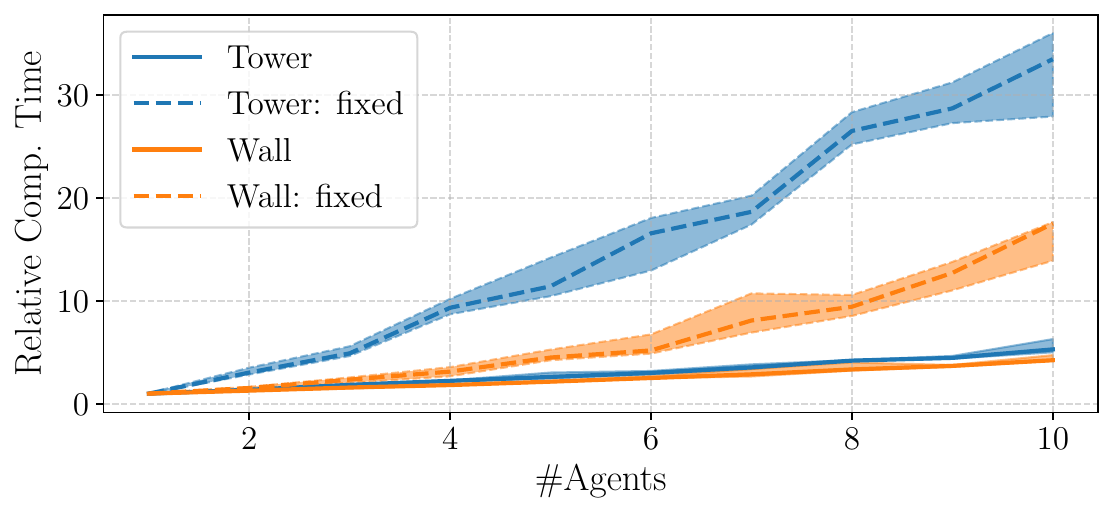} 
    \caption{
    The relative computation time for different numbers of agents for the tower and the wall scenario over 10 runs for our method and fixed-time sampling.
    The line is the median and the shaded area shows the 25 and 75 percentile, respectively.
    }
    \label{fig:ablation_rel_comp_time}
\end{figure}

\begin{figure}[t]
    \centering
    \includegraphics[width=.49\linewidth]{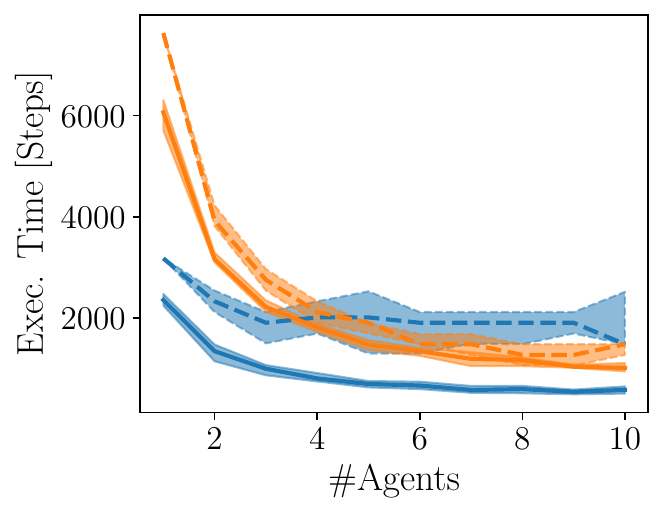} %
    \includegraphics[width=.49\linewidth]{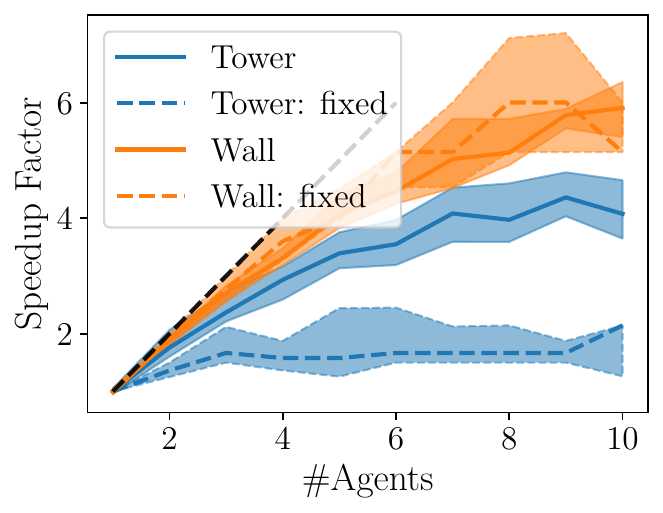} 
    \caption{
    Comparison of the \rb{execution time (left)}, and the speedup of the execution time when using multiple agents (right) for the tower and the wall models over 10 runs for our method and the fixed-time sampling.
    The speedup factor is the factor by which the execution time is sped up by using $m$ agents compared to only one agent.
    The line is the median and the shaded area is bounded by the 25 and 75 percentile.
    }
    \label{fig:ablation_rel_speedup}
\end{figure}


We compare the schedules generated by our approach, and the fixed-time-sampling approach for building the tower using 7 agents in \Cref{fig:ablation_schedule}:
It is clear that our method achieves better utilization in this example.
This is due to the fact that a single agent can temporarily block any other placement when building the tower.
\rb{This happens in the beginning, where blocks tend to be gripped from above, which  blocks placement of another block at the same time.
The later blocks need to be grabbed from the side to be placed, as the robots would not be able to place the parts that are higher up if they are grabbed at the top.
By coincidence, this enables parallelization for the later blocks for the fixed-time sampling in this example.}
Placing at a slightly different time would work but is not possible with this approach.

We show the same graphs as before, i.e. relative necessary computation time, and relative speedup in \cref{fig:ablation_rel_comp_time}, and \cref{fig:ablation_rel_speedup} respectively, showing that our method both achieves better speedup by using more robots, and better scaling with regard to computation time.
\Cref{tab:ablation_comp_time} shows the absolute computation times broken down to the different parts.
While we see that on the Wall-example, a similar speedup can be achieved due to the large space that is available, this is not the case in the Tower-example.
The main bottleneck in the fixed-time-sampling is the keyframe generation, which becomes much harder if the space is highly congested, and thus needs more restarts, which leads to a much longer computation time, and much worse scaling of the computation time with the number of agents we plan for.
%

\begin{table}[t]
\begin{center}
    \scriptsize
    \renewcommand\arraystretch{1.15}
    \caption{Median  computation times over 10 runs for the \textit{fixed-time} sampling method with number of agents $m$.
    The total time additionally contains e.g. pre-processing of the model, deciding on action-sequences etc.
    \rb{The super-{} and subscripts are the difference to the 25, and 75 percentile.}
    } 
    \begin{tabular}{p{10mm}p{2mm}|@{\hskip 1mm}
            S[table-format=3.1]@{\hskip 2mm}S[table-format=3.1]@{\hskip 2mm}S[table-format=3.1]@{\hskip 1mm}|@{\hskip 2mm}
            S[table-format=3.1]@{\hskip 6mm}}
    \toprule
    & & \multicolumn{4}{c}{Time [s]} \\
     & $m$ & \text{Keyframes-opt} & \text{Path-planning} & \text{Postprocessing} & \text{Total} \\
    \midrule
    \multirow{3}{*}{\shortstack[l]{Tower\\(15 obj.)}} 
                & 1 & 8.8\stackanchor{\scalebox{.6}{$+0.7$}}{\scalebox{.6}{$-0.6$}} & 1.0\stackanchor{\scalebox{.6}{$+0.1$}}{\scalebox{.6}{$-0.1$}} & 6.0\stackanchor{\scalebox{.6}{$+0.1$}}{\scalebox{.6}{$-0.1$}} & 17.3\stackanchor{\scalebox{.6}{$+0.8$}}{\scalebox{.6}{$-0.4$}} \\
                & 5 & 27.5\stackanchor{\scalebox{.6}{$+69.6$}}{\scalebox{.6}{$-8.2$}} & 6.4\stackanchor{\scalebox{.6}{$+0.6$}}{\scalebox{.6}{$-1.7$}} & 16.3\stackanchor{\scalebox{.6}{$+7.5$}}{\scalebox{.6}{$-0.4$}} & 197.6\stackanchor{\scalebox{.6}{$+48.1$}}{\scalebox{.6}{$-16.6$}} \\
                & 10 & 43.4\stackanchor{\scalebox{.6}{$+103.9$}}{\scalebox{.6}{$-1.8$}} & 11.2\stackanchor{\scalebox{.6}{$+5.8$}}{\scalebox{.6}{$-1.0$}} & 27.2\stackanchor{\scalebox{.6}{$+16.0$}}{\scalebox{.6}{$-0.1$}} & 579.1\stackanchor{\scalebox{.6}{$+43.8$}}{\scalebox{.6}{$-96.4$}} \\

    \midrule
    \multirow{3}{*}{\shortstack[l]{Wall\\ (36 obj.)}} 
                & 1 & 28.9\stackanchor{\scalebox{.6}{$+5.0$}}{\scalebox{.6}{$-1.9$}} & 4.9\stackanchor{\scalebox{.6}{$+0.1$}}{\scalebox{.6}{$-0.3$}} & 30.6\stackanchor{\scalebox{.6}{$+0.2$}}{\scalebox{.6}{$-0.2$}} & 70.6\stackanchor{\scalebox{.6}{$+3.8$}}{\scalebox{.6}{$-1.6$}} \\
                & 5 & 134.3\stackanchor{\scalebox{.6}{$+16.3$}}{\scalebox{.6}{$-18.3$}} & 24.2\stackanchor{\scalebox{.6}{$+4.1$}}{\scalebox{.6}{$-1.5$}} & 61.2\stackanchor{\scalebox{.6}{$+4.2$}}{\scalebox{.6}{$-0.3$}} & 317.8\stackanchor{\scalebox{.6}{$+54.5$}}{\scalebox{.6}{$-18.6$}} \\
                & 10 & 207.8\stackanchor{\scalebox{.6}{$+16.5$}}{\scalebox{.6}{$-24.0$}} & 43.7\stackanchor{\scalebox{.6}{$+21.4$}}{\scalebox{.6}{$-1.1$}} & 91.4\stackanchor{\scalebox{.6}{$+0.3$}}{\scalebox{.6}{$-1.6$}} & 1235.8\stackanchor{\scalebox{.6}{$+9.5$}}{\scalebox{.6}{$-254.8$}} \\

    \bottomrule
    \end{tabular}
    \label{tab:ablation_comp_time}
\end{center}
\end{table}

\subsection{Demonstrations}
\subsubsection{Long horizon assembly}
We demonstrate our algorithm on two long horizon-construction tasks using the mobile manipulators, and modeling manipulation constraints as gripping-by-touch:
\begin{itemize}
    \item The well, using 6 agents: Computation time 14.1 min, execution time 2150 steps.
    \item The pavilion, using 8 agents: Computation time 43.2 min, execution time 3867 steps.
\end{itemize}
We show a schedule for the assembly of the pavilion using 8 agents in~\cref{fig:schedule} to showcase the complexities in coordinating the robot movements.


\subsubsection{Handover scenario}
\begin{figure}[t]
    \centering
    \includegraphics[width=.97\linewidth]{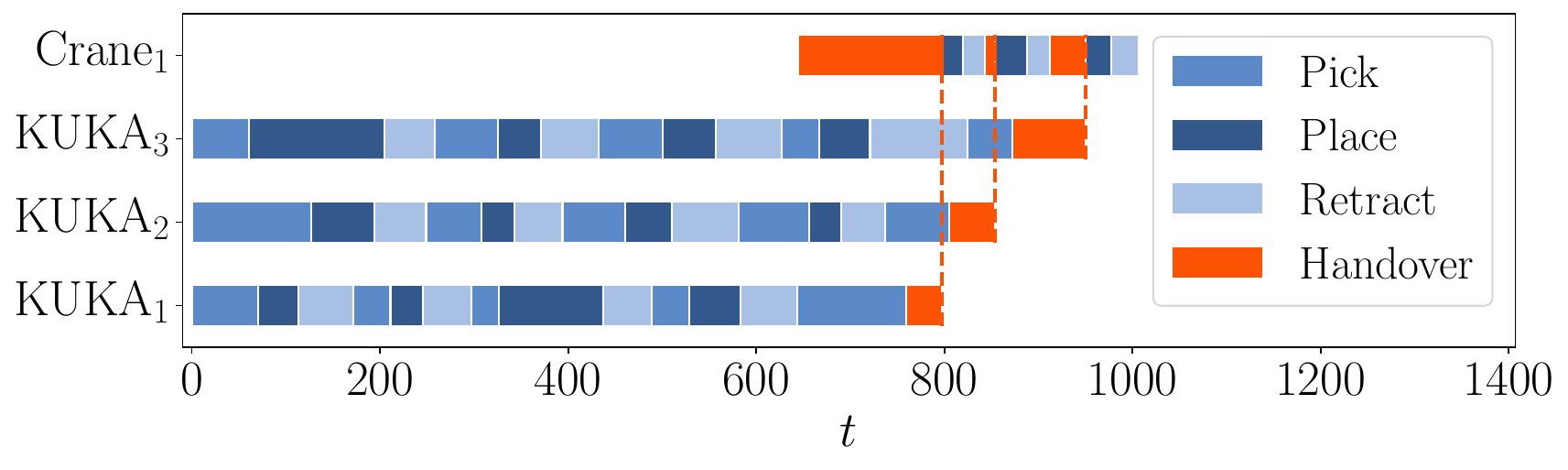} 
    \caption{
    Illustration of a schedule for the Tower-example using three KUKA-arms on mobile bases and a tower crane, therefore requiring robot-robot interaction. 
    We specifically highlight the times at which constraints are active for two robots (\textit{handover}), and which agents are affected.
    }
    \label{fig:handover_schedule}
\end{figure}

\begin{figure*}
\centering
    \begin{subfigure}[t]{.249\textwidth}
        \centering
        \includegraphics[height=0.8\linewidth,width=.97\linewidth]{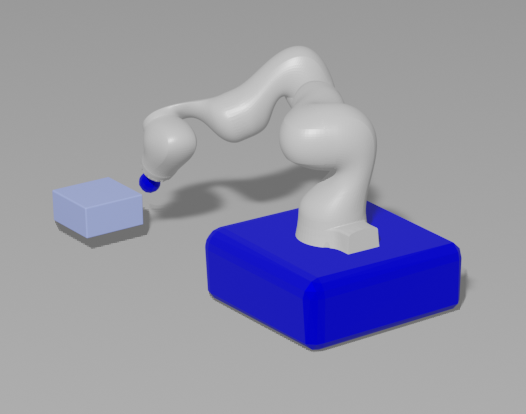} 
        \caption{}
    \end{subfigure}%
    \begin{subfigure}[t]{.249\textwidth}
        \centering
        \includegraphics[height=0.8\linewidth,width=.97\linewidth]{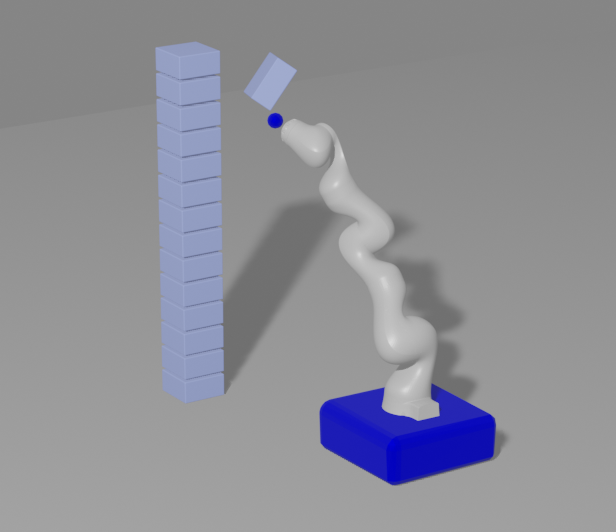}
        \caption{}
    \end{subfigure}
    \begin{subfigure}[t]{.249\textwidth}
        \centering
        \includegraphics[height=0.8\linewidth,width=.97\linewidth]{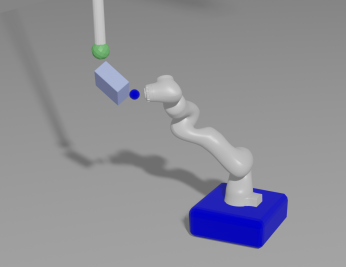}
        \caption{}
    \end{subfigure}%
    \begin{subfigure}[t]{.249\textwidth}
        \centering
        \includegraphics[height=0.8\linewidth,width=.97\linewidth]{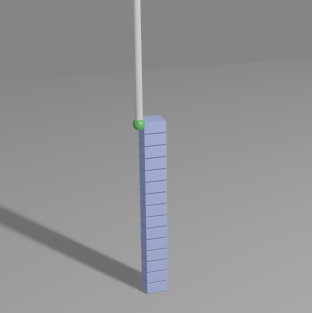}
        \caption{}
    \end{subfigure}
    \caption{Illustration of a handover scenario with a KUKA-arm and a crane. \textbf{(a)} KUKA-arm picks up the object. \textbf{(b)} The placement position is not reachable. \textbf{(c)} A handover to the crane is executed. \textbf{(d)} The crane places the object onto tower.}
    \label{fig:handover}
\end{figure*}

We consider the scenario of the tower again, but this time with three mobile bases with KUKA-arms on top.
Manipulation is modeled as gripping-by-touch.
Since the KUKA-arms are unable to reach the top of the tower, we add a tower crane.
However, the crane is unable to reach the pieces on the floor.
Therefore, \textit{handover} sequences are necessary to place the last 3 parts.
This scenario demonstrates the ability of our framework to handle and coordinate robots with different capabilities and explore various possible task sequences to fulfill a task.
The resulting schedule, and some frames from the process can be seen in \cref{fig:handover_schedule} and \cref{fig:handover}, respectively.

\rb{
\subsubsection{Real robot experiments}
We demonstrate an experiment with two robotic arms with Robotiq grippers as end-effectors.
We model the manipulation constraints of the robot as a two finger gripper.
The goal is to stack 6 boxes using the two arms.
}

\begin{figure*}
\centering
    \begin{subfigure}[t]{.2\textwidth}
        \centering
        \includegraphics[trim={1.5cm 0.3cm 1.5cm 0.3cm},clip,width=.97\linewidth]{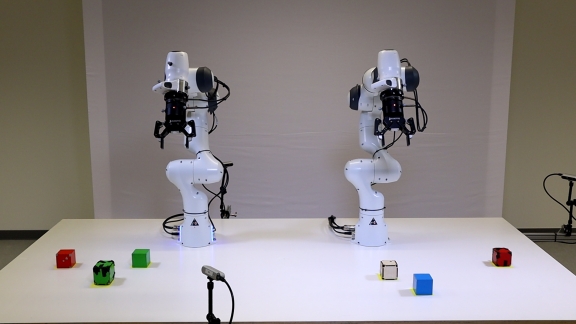} 
        \caption{}
    \end{subfigure}%
    \begin{subfigure}[t]{.2\textwidth}
        \centering
        \includegraphics[trim={1.5cm 0.3cm 1.5cm 0.3cm},clip,width=.97\linewidth]{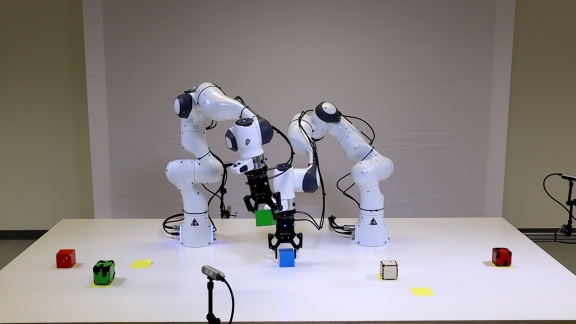}
        \caption{}
    \end{subfigure}
    \begin{subfigure}[t]{.2\textwidth}
        \centering
        \includegraphics[trim={1.5cm 0.3cm 1.5cm 0.3cm},clip,width=.97\linewidth]{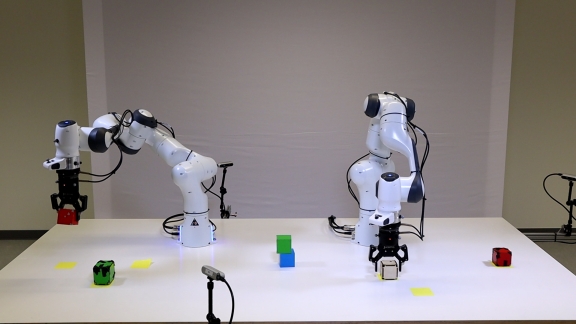}
        \caption{}
    \end{subfigure}%
    \begin{subfigure}[t]{.2\textwidth}
        \centering
        \includegraphics[trim={1.5cm 0.3cm 1.5cm 0.3cm},clip,width=.97\linewidth]{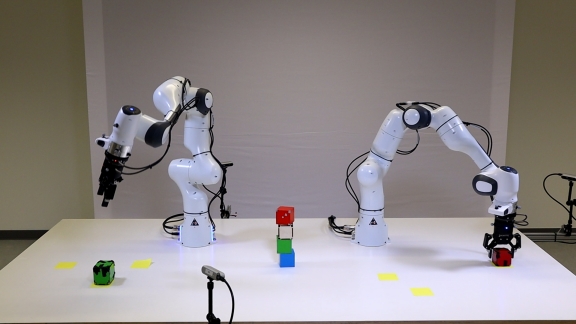}
        \caption{}
    \end{subfigure}%
    \begin{subfigure}[t]{.2\textwidth}
        \centering
        \includegraphics[trim={1.5cm 0.3cm 1.5cm 0.3cm},clip,width=.97\linewidth]{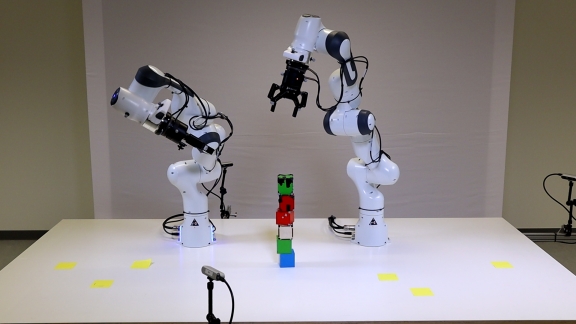}
        \caption{}
    \end{subfigure}
    \caption{
    \rb{Snapshots of the robot experiment, where two Panda arms with two-finger gripper end-effectors are stacking six boxes.}
    }
    \label{fig:pandas}
\end{figure*}

\rb{
We execute the trajectory open-loop; we thus rely on an accurate model in the simulation, such that the trajectory can be executed without adaptions.
\Cref{fig:pandas} shows a sequence of images from the two robots placing the blocks.
While it is visible that the boxes are not perfectly aligned, it is clear that the algorithm succeeds in effectively coordinating the robots.
This showcases that it is possible to account for the constraints arising in real robot experiments and that the plans and trajectories generated by our algorithm can be executed on a set of real robots.
}


\section{Discussion \& Limitations}
Multi-robot assembly is a complex problem which consists of multiple NP-hard subproblems \cite{Reif1994}. 
The method development in this paper is application-driven, aiming to push towards an efficient solution strategy that scales well to challenging scenarios.
There are parts of the approach that were not considered in this work, which need to be tackled when deploying this framework to real construction scenarios.

\rb{
At the moment, the algorithm is too slow to allow for online-replanning. 
As visible from the experiments, the keyframe-computation take the majority of the time.
Smarter generation of the keyframes, e.g., through online learning of initializations for the optimizer, is one way to decrease the time \cite{ortiz2022structured, driess2021learning}.
Reducing the number of sampled goal states is another possibility to scale down the necessary time.
For speeding up the planning, the biggest possible improvement is the usage of multiquery planning.
Applying multiquery path planning in a dynamic environment is not straightforward however, and needs to be considered as future work.
}

\rb{
While we showed that the paths can be executed by real robots directly in a controlled environment, in case uncertainty is present, this might be more challenging.
Hence, the planning done in this framework might need to account for uncertainty in the execution, both in space, and in time.
One possibility to do so would be to reserve a `safe corridor' for an agent, through which no other agent travels for a given time-window.
Another one would be to, e.g., maximize clearance during the path planning, and not only to minimize time.
}

\subsection{Discussion of Theoretical Properties}\label{sec:comp}
We briefly discuss the properties of the algorithm, and necessary changes to achieve completeness.


\paragraph{Selection of Subproblems}
The priorization of objects and robots in combination with the backtracking can be seen as depth first search.
This means that every possible assignment will be chosen at some point. 
The methods we use for path-planning can not prove an infeasibility, which is why we need to continue exploring the `infeasible' nodes.
\paragraph{Choice of action-sequence} 
We are using a depth-first search over the space defined by a first-order logic-language to compute the available action sequences.
Under the assumption that each object can be rearranged to its goal position within a finite number of actions, this implies that we will find a feasible action sequence, if one exists.
\paragraph{Path-planning}
Path-planning in our algorithm works sequentially through an action-sequence, and plans each part of the sequence using an RRT.
This means that each segment alone is asymptotically complete in the limit, assuming that the keyframe-sampling is uniformly covering the solution manifold and that we return to a subproblem infinite number of times.
It might be possible that we find a path in one part of the action sequence which does not have a corresponding feasible path in the following action, and thus label the sequence as `infeasible'.
The occurrence of this is greatly reduced by jointly sampling all keyframes.
Continuously re-expanding the `infeasible' nodes covers this case.

We finally note that we need to make sure in the algorithm that we do not make subsequent placements impossible by fixing paths that do not allow for a feasible subsequent path, i.e., do not block \emph{pick} or \emph{place} configurations with `resting' agents.
\section{Conclusion}

We presented a planning system to solve long-horizon multi-robot construction assembly problems that integrates several novel components.
The approach strongly exploits the factorizations of multi-agent construction assembly problems, by solving simpler subproblems involving only a subset of agents to plan for and a single object that has to be placed at its goal location.
These solutions to the separate subproblems are then used to construct a feasible solution to the overall problem.
To solve the limited-horizon subproblems, we combine sampling based path planning with joint mode-switch optimization to solve for manipulation constraints, and proposed novel methods to find time-embeddings for planned tasks.
Path planning between keyframes amidst other moving, previously planned, objects and robots is achieved using a novel bi-directional RRT-planner in space-time.

We demonstrate that our approach scales well to many robots and many objects on a variety of construction tasks.
We provide both qualitative and quantitative analysis of the results.
Compared to planning task assignments at fixed times, our time-embedding leads to better utilization of the robots and hence lower execution time to achieve the task, as well as lower computation times for planning the movements.
Finally, we demonstrated the approach in a real robot experiment.
\rb{
The robotic experiments show that it is possible to execute the paths that our approach generates.
}

The approach exploits decompositionality and greedily selects the next sub-tasks. 
This is successful for our application scenarios, but compromises global optimality for efficient planning and execution times, as is crucial to make multi-robot planning work.

We want to push the approach to demonstration on real construction scenarios.
In this setting, more realism in the model description is required, including exact physical constraints on static stability.

\section{Acknowledgement}
The authors thank Christoph Schlopschnat for the model of the wooden pavilion.

\bibliographystyle{IEEEtran}
\bibliography{IEEEabrv,bib/general.bib}
\end{document}